\documentclass[sigconf]{acmart}
\pdfoutput=1

\usepackage{xcolor}
\usepackage{enumitem}
\usepackage{mdwlist}
\usepackage[font={small,bf},labelfont=bf,format=plain]{caption}
\usepackage{amsmath,amssymb,amsthm}
\usepackage{subcaption}
\usepackage[labelfont=bf,format=plain]{caption}
\usepackage{xspace}
\usepackage{wrapfig}
\usepackage{amsthm}
\usepackage{mathtools}
\usepackage{algorithm}
\usepackage{algpseudocode}
\algtext*{EndProcedure}
\algtext*{EndFor}
\algtext*{EndWhile}
\algtext*{EndIf}
\usepackage{colortbl}
\usepackage{multicol}
\usepackage{multirow}

\newcolumntype{H}{>{\setbox0=\hbox\bgroup}c<{\egroup}@{}}

\newcommand*\mystrut[1]{\vrule width0pt height0pt depth#1\relax}

\makeatletter
\def\smallunderbrace#1{\mathop{\vtop{\m@th\ialign{##\crcr
   $\hfil\displaystyle{#1}\hfil$\crcr
   \noalign{\kern3\p@\nointerlineskip}
   \tiny\upbracefill\crcr\noalign{\kern3\p@}}}}\limits}
\makeatother

\newcolumntype{C}[1]{>{\centering\let\newline\\\arraybackslash\hspace{0pt}}m{#1}}

\newtheorem{problem}{Problem}

\definecolor{Green}{rgb}{0.34,0.52,0.14}
\definecolor{Gray}{gray}{0.95}

\newcommand{\KGgold}{\hat{\mathcal{G}}}
\newcommand{\nodesGold}{\hat{\nodes}}
\newcommand{\edgesGold}{\hat{\edges}}
\newcommand{\nodeLabelsGold}{\hat{\mathcal{L}}_\nodes}
\newcommand{\relationsGold}{\hat{\mathcal{L}}_\edges}
\newcommand{\phiGold}{\hat{\phi}}
\newcommand{\KGLong}{(\nodes, \edges, \nodeLabels, \relations, \phi)}

\newcommand{\triple}{t}
\newcommand{\tripleLong}{(s, p, o)}
\newcommand{\warAndPeace}{\texttt{\small War \& Peace}}
\newcommand{\frankenstein}{\texttt{\small Frankenstein\xspace}}
\newcommand{\tolstoy}{(\texttt{\small War \& Peace, writtenBy, Leo Tolstoy})}

\DeclareMathOperator*{\argmin}{arg\,min}

\def\model{\ensuremath{\mathit M}}

\newcommand{\Model}{M}
\newcommand{\Models}{\mathcal{M}}
\newcommand{\DB}{\mathcal{D}}
\newcommand{\mdl}{L}

\newcommand{\NELL}{{\texttt{NELL}}}
\newcommand{\DBpedia}{{\texttt{DBpedia}}}
\newcommand{\Yago}{{\texttt{Yago}}}

\newcommand{\method}{\textsc{KGist}\xspace}

\newcommand{\RefTwo}{\textbf{Rm}}
\newcommand{\RefThree}{\textbf{Rn}}
\newcommand{\RefTwoShort}{{m}}
\newcommand{\RefThreeShort}{{n}}

\newcommand{\methodTwo}{\method{}+\RefTwoShort} 
\newcommand{\methodThree}{\method{}+\RefThreeShort} 

\newcommand{\kgistNull}{\textsc{KGist}\xspace}
\newcommand{\kgistFreq}{\textsc{KGist\_Freq}\xspace}
\newcommand{\freq}{Freq}
\newcommand{\cover}{Coverage}
\newcommand{\sdv}{SDValidate}
\newcommand{\transe}{TransE}
\newcommand{\complex}{ComplEx}
\newcommand{\amie}{AMIE+}
\newcommand{\amieC}{AMIE+C}

\newcommand{\naturalEncoding}{L_\mathbb{N}}
\newcommand{\modelSpace}{\mathcal{M}}
\newcommand{\emptyModel}{\model_0}

\newcommand{\goldOptimalModel}{M^*}
\newcommand{\adjacencyTensor}{\mathbf{A}}
\newcommand{\explainedEdges}{\mathbf{A}_M}
\newcommand{\labMatrix}{\mathbf{L}}
\newcommand{\explainedLabels}{\mathbf{L}_M}
\newcommand{\candidates}{\mathcal{C}}
\newcommand{\powerset}{\mathcal{P}}

\newcommand{\graph}{G}
\newcommand{\nodes}{\mathcal{V}}
\newcommand{\edges}{\mathcal{E}}
\newcommand{\numNodes}{|\nodes|}
\newcommand{\numEdges}{|\edges|}
\newcommand{\labels}{\phi}
\newcommand{\nodeLabels}{\mathcal{L_V}}
\newcommand{\relations}{\mathcal{L_E}}
\newcommand{\numRelations}{|\mathcal{L_E}|}
\newcommand{\numNodeLabels}{|\mathcal{L_V}|}

\newcommand{\ruleGraph}{g}
\newcommand{\ruleChild}{\hat{\ruleGraph}}
\newcommand{\dirOut}{\rightarrow}
\newcommand{\dirIn}{\leftarrow}
\newcommand{\predicate}{p}
\newcommand{\dir}{\delta}
\newcommand{\ruleRoot}{\mathcal{L}_\ruleGraph}
\newcommand{\RootLabelSet}{\mathcal{\varPhi}_\ruleGraph}
\newcommand{\ruleChildren}{\chi_\ruleGraph}
\newcommand{\rules}{{\model}}
\newcommand{\assertion}{a}

\newcommand{\traversalStart}{s_{a_g}}
\newcommand{\assertions}{\mathcal{A}}
\newcommand{\correctAssertions}{\assertions_c}

\newcommand{\exceptions}{\assertions_\xi}

\newcommand{\anom}{\eta}

\newcommand{\qOne}{\textbf{Q1}}
\newcommand{\qTwo}{\textbf{Q2}}
\newcommand{\qThree}{\textbf{Q3}}
\newcommand{\qFour}{\textbf{Q4}}

\newcommand{\anomOne}{\textbf{A1}}
\newcommand{\anomTwo}{\textbf{A2}}
\newcommand{\anomThree}{\textbf{A3}}
\newcommand{\anomFour}{\textbf{A4}}
\newcommand{\samplingProb}{q}

\newcommand{\negativeEdgeError}{\mathbf{A}^-}
\newcommand{\negativeEdgeErrorElement}{\mathbf{A}_{s,o,p}^-}
\newcommand{\negativeLabelError}{\mathbf{L}^-}

\newcommand{\order}{O}

\copyrightyear{2020}
\acmYear{2020}
\setcopyright{iw3c2w3}
\acmConference[WWW '20]{Proceedings of The Web Conference 2020}{April 20--24, 2020}{Taipei, Taiwan}
\acmBooktitle{Proceedings of The Web Conference 2020 (WWW '20), April 20--24, 2020, Taipei, Taiwan}
\acmPrice{}
\acmDOI{10.1145/3366423.3380189}
\acmISBN{978-1-4503-7023-3/20/04}

\begin{document}
\title[{What is Normal, What is Strange, and What is Missing in a KG: Unified Characterization via Inductive Summarization}]{What is Normal, What is Strange, and What is Missing in a Knowledge Graph: Unified Characterization via Inductive Summarization}

\author{Caleb Belth}
\affiliation{
  \institution{University of Michigan}
}
\email{cbelth@umich.edu}

\author{Xinyi Zheng}
\affiliation{
  \institution{University of Michigan}
}
\email{zxycarol@umich.edu}

\author{Jilles Vreeken}
\affiliation{
  \institution{CISPA Helmholtz Center for Information Security}
}
\email{jv@cispa.saarland}

\author{Danai Koutra}
\affiliation{
  \institution{University of Michigan}
}
\email{dkoutra@umich.edu}

\begin{abstract}
Knowledge graphs (KGs) store highly heterogeneous information about the world in the structure of a graph, and are useful for tasks such as question answering and reasoning. However, they often contain errors and are missing information. Vibrant research in KG refinement has worked to resolve these issues, tailoring techniques to either detect specific types of errors or complete a KG.

In this work, we introduce a \textit{unified solution} to KG characterization by formulating the problem as \emph{unsupervised KG summarization} with a set of inductive, \textit{soft rules}, which describe what is \emph{normal} in a KG, and thus can be used to identify what is \emph{abnormal}, whether it be strange or missing. Unlike first-order logic rules, our rules are labeled, rooted graphs, i.e., patterns that describe the expected neighborhood around a (seen or unseen) node, based on its type, and information in the KG. Stepping away from the traditional support/confidence-based rule mining techniques, we propose \method, \emph{Knowledge Graph Inductive SummarizaTion}, which learns a summary of inductive rules that best compress the KG according to the Minimum Description Length principle---a formulation that we are the first to use in the context of KG rule mining. We apply our rules to three large KGs (\NELL{}, \DBpedia{}, and \Yago{}), and tasks such as compression, various types of error detection, and identification of incomplete information. We show that \method outperforms task-specific, supervised and unsupervised baselines in error detection and incompleteness identification, (identifying the location of up to 93\% of missing entities---over 10\% more than baselines), while also being efficient for large knowledge graphs.
\end{abstract}

\maketitle

\section{Introduction}
\label{sec:intro}

\enlargethispage{\baselineskip}

\begin{figure}[t!]
    \centering
    \includegraphics[width=\columnwidth]{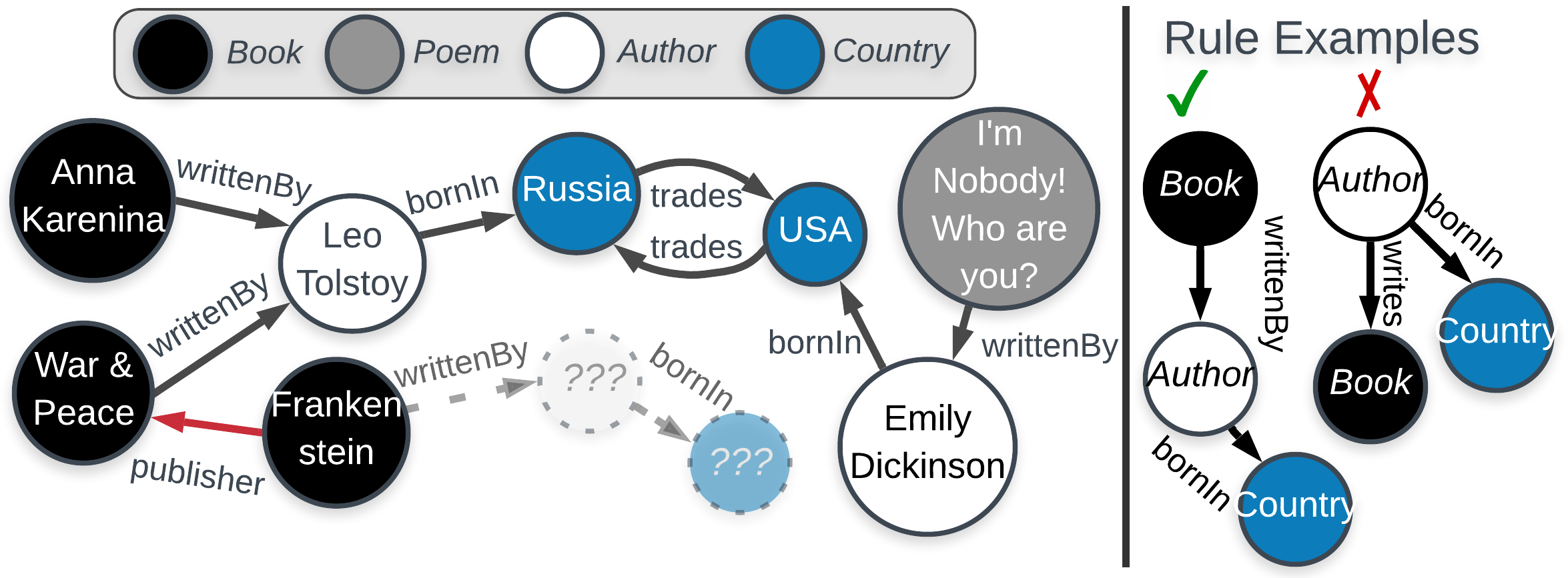}
    \vspace{-0.1cm}
    \caption{
    \method{} summarizes a KG (left) by finding patterns that can be interpreted as rules (right). For instance, the rule that \texttt{\footnotesize books are written by authors, who are born in countries}, which holds in two out of three cases in this KG (\texttt{\footnotesize Frankenstein} is missing an author), correctly describes books in general. However, the opposite pattern does not: while \texttt{\footnotesize Leo Tolstoy} writes books, \texttt{\footnotesize Emily Dickinson} writes poems. The summary of rules characterizes what is \emph{normal} in a KG, while simultaneously revealing what is \emph{strange} and \emph{missing}, such as the erroneous and missing edges around \texttt{\footnotesize Frankenstein}.
    }
    \label{fig:intro}
    \vspace{-0.25cm}
\end{figure}

Knowledge graphs (KGs), such as \NELL{} \cite{carlson2010toward}, \DBpedia{} \cite{auer2007dbpedia},
and \Yago{} \cite{suchanek2007yago}, store collections of entities and relations among those entities (Fig.~\ref{fig:intro}), and are often used for tasks such as question answering, powering virtual assistants, reasoning, and fact checking \cite{huang2019knowledge, bhutani2019learning, nickel2015review, shiralkar2017finding}. Many KGs 
encode encyclopedic information, i.e., facts about the world, and are, to a large degree, automatically built \cite{nickel2015review}. As a result,
they contain many types of errors, and are missing edges, nodes, and labels. 
This has led to a significant amount of research on KG refinement, resulting in {\em task-specific} methods that {\em either} identify erroneous facts 
{\em or} add new ones~\cite{paulheim2017knowledge}. 
While the accuracy of KG tasks may be improved by refinement, KGs grow to the order of millions or billions of edges,
making KGs more inaccessible to users \cite{huang2019knowledge}, and tasks over them more computationally difficult \cite{nickel2015review}.

As refinement helps address accuracy issues, graph summarization \cite{liu2018graph} can help address
KG size issues by
describing a graph with simple and concise patterns.
However, KG-specific summarization \cite{zneika2019quality} focuses mostly on query- or search-related summaries~\cite{song2018mining,wu2013summarizing,safavi2019personalized}, while most general-graph summarization work is designed for purposes other than KG refinement, and aims to compress a graph by grouping together similarly linked and similarly labeled nodes. 
These summaries would only cluster existing information in a KG,
but encyclopedic KGs will always be missing facts (since the world's information is unbounded).

Thus, we introduce the problem of \emph{inductive KG summarization}, in which,
given a knowledge graph $\graph$, we seek to find a concise and interpretable summary of $\graph$ with \emph{inductive} rules that can \emph{generalize} to the parts of the world not captured by $\graph$. With this characterization of the norm, we can also identify \emph{what is strange} in and \emph{what is missing} from $\graph$: the parts of the graph that violate the rules or remain unexplained by the summary. These strange parts of the graph may be genuine exceptions, errors, or missing information.
To solve the problem, we propose \method{}, an information-theoretic approach that serves as a \textit{unified} solution to summarization and various KG refinement tasks, which have traditionally been viewed independently.

{Our main contributions are summarized as follows:}
\begin{itemize}[wide]
    \item \textbf{Problem Formulation}. Rather than targeting a specific refinement task (e.g., link prediction), we unify various refinement tasks by joining the problems of refinement and unsupervised summarization, and introduce the notion of inductive summarization with soft rules that plausibly generalize beyond the KG. \S~\ref{sec:model}
    \item \textbf{Expressive rules.} While current methods (\S~\ref{sec:related}) learn first-order logic rules that have single-element consequences, which predict single edges, our rules are labeled, rooted graphs that are recursively defined, allowing them to describe {\em arbitrary} graph structure around a node (i.e., they can have complex consequences). Our formulation of rules takes a step towards treating knowledge \emph{graphs} as graphs---something often overlooked in KG refinement \cite{paulheim2017knowledge}. \S~\ref{sec:model}
    \item \textbf{MDL-based approach.} We introduce \method, an unsupervised, information-theoretic approach  
    that identifies rules via the Minimum Description Length (MDL) principle~\cite{Rissanen85Minimum}, going beyond the support/confidence framework of prior work.  \S~\ref{sec:algorithm}
    \item \textbf{Experiments on real KGs.} We perform extensive experiments on large KGs (\NELL{}, \DBpedia{}, \Yago{}), and
    diverse
    tasks, including compression, various types of error detection, and identifying the absence of nodes.
    We show that \method{} learns orders of magnitude fewer rules than current methods, allowing \method{} to be efficient and effective at diverse tasks.
   \method identifies the location of 76-93\% of missing entities---over 10\% more than baselines. \S~\ref{sec:eval}

\end{itemize}
Our code and data are available at {\href{https://github.com/GemsLab/KGist}{https://github.com/GemsLab/KGist}}.

\section{Related Work}
\label{sec:related}

\subsection{Knowledge Graph Refinement} 
KG refinement attempts to resolve erroneous or missing information~\cite{paulheim2017knowledge, paulheim2014improving}.
Next, we discuss the three most relevant categories of refinement techniques (although other methods exist, such as crowd-sourcing-based methods \cite{jiang2018knowledge}).

\subsubsection{Rule-mining-based Refinement.} 
These approaches are reminiscent of association rule mining \cite{agrawal1993mining}.
AMIE~\cite{galarraga2013amie} introduces an altered
\emph{confidence} metric based on the 
partial completeness assumption,
according to which, if a particular relationship of an entity is known, then all relationships of that type for that entity are known (as opposed to the \emph{open-world assumption}, which
assumes that an absent relationship could either be missing or not hold in reality). \amie{} \cite{galarraga2015fast} is optimized to scale to larger KGs, and 
Tanon \emph{et al.} \cite{tanon2017completeness} seek to acquire and use counts of edges to measure the incompleteness of KGs. Other, non-rule-mining-based methods have also been proposed for measuring KG quality  \cite{rashid2019quality, jia2019triple}.
A supervised approach that augments \amie{}~\cite{galarraga2017predicting} takes example complete and incomplete assertions (e.g., crowd-sourced) as training data, and predicts completeness of predicate types observed during training.  
These works focus on refinement and find Horn rules on \emph{binary predicates}.
In contrast, we focus on summarization, and our rules can be applied to a \emph{node}, knowing only its type.
Also, we go beyond the support/confidence framework, which treats KGs as a table of transactions, and take a graph-theoretic view instead.
One work that \emph{does} take a graph-theoretic view learns rules in a bottom-up fashion by sampling paths from the KG, but the rules are constrained to be path-based Horn-rules~\cite{meilicke2019anytime}. Graph-Repairing Rules (GRRs) \cite{cheng2018rule} have also been proposed to target the specific problems of identifying incomplete, conflicting, and redundant information in graphs. They focus on simple graphs, whereas KGs contain multi-edges \cite{nickel2015review}, multiple labels per node (Tab~\ref{table:stats}), and self-loops.  GRRs were preceded by less expressive association rules with graph patterns \cite{fan2015association} and functional dependencies for graphs \cite{fan2016functional}. Rule-mining also has applications beyond KG refinement, such as recommender systems \cite{ma2019jointly}. Our rules could potentially be used in these scenarios, but we leave that for future work.

\subsubsection{Embedding-based Refinement.} KG embedding approaches seek to learn representations of nodes and relations in a latent space \cite{wang2017knowledge}, spanning from tensor factorization-based methods~\cite{nickel2011three, nickel2012factorizing} to translation-based methods such as \transe{}~\cite{bordes2013translating} and semantic matching models such as \complex{}~\cite{trouillon2016complex}. These works often perform link prediction, which is useful for completing relationships among entities, but only predicts links between entities already in the KG. In contrast, \method
can identify the \emph{absence} of entities from the KG.

\subsubsection{Hybrid Refinement.}
Recent refinement methods improve link prediction performance by iteratively applying rule mining and learning embeddings. For instance, pre-trained embeddings have been used to more accurately measure the quality of candidate rules~\cite{ho2018rule}. In~\cite{zhang2019iteratively}, facts inferred from rules improve embeddings of sparse entities, and in turn embeddings improve the efficiency of rule mining. Unlike these works, we focus on unifying different refinement tasks, going beyond link prediction.

\subsection{Graph Summarization}
\label{subsec:graph-sum}
Graph summarization seeks to succinctly describe a large graph in a smaller representation either in the original or a latent space~\cite{liu2018graph,JinRKKRK19}.
Much of the work on \emph{knowledge graph} summarization has focused on query-related summaries, such as query answer-graph summaries~\cite{wu2013summarizing},
patterns that can be used as query views to improve KG search~\cite{song2018mining,fan2014answering},
and sparse, personalized KG summaries---based on historical user queries---for use on personal, resource-constrained devices~\cite{safavi2019personalized}. While our summaries could conceptually be used for query-related problems, we focus on the problem of characterizing what is normal, strange, and missing in a KG. We also construct summaries with patterns that generalize, which is not considered by \cite{song2018mining}.
Similar to summarization, Boded \emph{et al.} \cite{Bobed2019DatadrivenAO} use MDL to assess KG evolution, but they do not target refinement. Beyond KGs, MDL has been used to summarize static and temporal graphs via structures, such as cliques, stars, and chains~\cite{koutra2014vog,shah2015timecrunch,NavlakhaRS08bounded,Goebl1TBP6}, or frequent subgraphs \cite{noble2003graph} (also studied from the perspective of subgraph support \cite{elseidy2014grami}).
Unlike these works, we learn \textit{inductive} summaries of recursively defined \textit{rules} or rooted graphs, which incorporate both the KG structure and semantics, and can be used for graph refinement.

\section{Inductive Summarization: Model}
\label{sec:model}
In this section we describe our proposed MDL formulation for inductive summarization of knowledge graphs, after introducing some preliminary definitions. We list the most frequently used symbols in Table~\ref{tab:Symbols}, along with their definitions.

\subsection{Preliminaries}

\subsubsection{\bf Knowledge Graph (KG) $G$} A KG is a labeled, directed graph $\graph = \KGLong$, consisting of a set of nodes or entities  $\nodes$, a set of relationship types $\relations$, a set of edges or triples $\edges \in \nodes \times \relations \times \nodes$, a set of node labels $\nodeLabels$, and a function $\labels : \nodes \rightarrow \powerset(\nodeLabels)$ mapping nodes to their labels, the set of which we call the node's \emph{type}. 
We give an example KG in Fig.~\ref{fig:intro}.
An edge or triple $\triple = \tripleLong$ connects the \textit{subject} and \textit{object} nodes $s, o \in \nodes$ via a relationship type (\textit{predicate}) $p \in \relations$.
An example is {\small \tolstoy}.
Triples encode a unit of information or fact, semantically {about} the subject.
Since a pair of nodes may have multiple edges between them, we represent the connectivity of $\graph$ with a $\numNodes \times \numNodes \times \numRelations$ adjacency tensor $\adjacencyTensor$.
Similarly, we store the label information in an $\numNodeLabels \times \numNodes$ binary label matrix, $\labMatrix$.
\subsubsection{\bf Ideal Knowledge Graph $\KGgold$} An \textit{ideal} knowledge graph $\KGgold(\nodesGold, \edgesGold, \nodeLabelsGold, \relationsGold, \phiGold)$ contains all the correct facts in the world and no incorrect ones, i.e., $\tripleLong \in \edgesGold$ if and only if the fact holds in reality. An ideal KG is only a conceptual aid, and does not exist, since KGs have errors and missing information.

\subsubsection{\bf Model $\model$ of a KG} \label{subsec:model-def} A model $\model$ of a KG is a \textit{set of inductive rules}, which describe its facts (see formal definition in \S~\ref{subsec:rule}). In \S~\ref{sec:mdl}, we will explain a model in the context of our work.

\begin{table}[!t]
{\small
\caption{Description of major symbols.}
\vspace{-0.35cm}
\label{tab:Symbols}
\centering
  \begin{tabular}{ll}
  \toprule
     \textbf{Notation} & \textbf{Description}  \\ \midrule
     $\graph(\nodes,\edges)$ & knowledge graph\\
     $\adjacencyTensor$, $\labMatrix$ & binary adjacency tensor and label matrix of $\graph$, resp. \\
     $\model, \emptyModel$ & model or set of rules, and the empty model, resp.\\ 
     $L(.)$ & \# of bits to transmit an object (e.g., a graph or rule)\\
     $\ruleGraph$ & rule in the form of a graph pattern \\
     $\assertions^{(\ruleGraph)}$, $\correctAssertions^{(\ruleGraph)}$, $\exceptions^{(\ruleGraph)}$ & assertions, correct assertions, exceptions of $\ruleGraph$, resp.\\
     $|.|$ & set cardinality and number of 1s in tensor/matrix\\
     \bottomrule
    \end{tabular}
    }
\end{table}

\subsubsection{{\bf Rule} $\ruleGraph$} \label{subsec:rule} 
\begin{figure}[t!]
    \centering
    \includegraphics[width=.87\columnwidth]{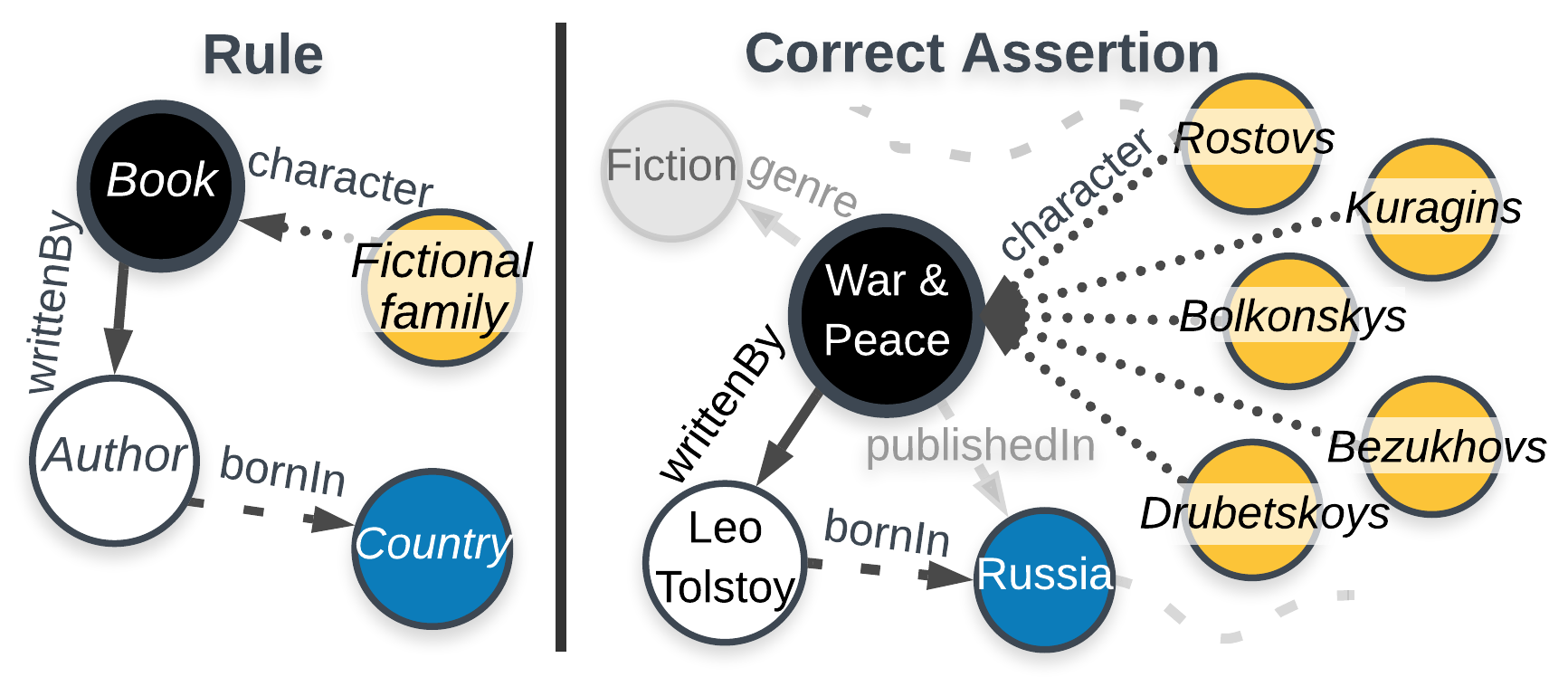}
    \vspace{-0.4cm}
    \caption{An example rule and one of its correct assertions.
    The correct assertion is a traversal starting at \texttt{\footnotesize War \& Peace} because it is a \texttt{\footnotesize Book} (root), and following the rule's syntax to induce a subgraph (line styles denote edge types).
    For instance, the first child of the rule lexicographically is \texttt{\footnotesize (character, $\dirIn$, (\{Fictional Family\}, $\emptyset$))}, which would be traversed recursively if it were not a leaf rule. This part of the rule asserts that \texttt{\footnotesize books} have one or more \texttt{\footnotesize Fictional Family characters}.
    During the traversal, \emph{every} neighboring node that matches the rule's syntax is traversed (e.g. all the fictional families are visited). Traversals from all \texttt{\footnotesize Book} nodes constitute $\assertions^{(\ruleGraph)}$. If a node lacks a neighbor asserted by the rule (e.g. if \texttt{\footnotesize Leo Tolstoy} had no \texttt{\footnotesize bornIn} edge), then it is an \emph{exception}.}
    \label{fig:rule-example}
    \vspace{-0.45cm}
\end{figure}

A rule $\ruleGraph \in \model$ is defined \textit{recursively} and \textit{compositionaly}. 
Specifically, rule $\ruleGraph = (\ruleRoot, \ruleChildren)$ is a rooted, directed graph, with a subset of node labels $\ruleRoot \subseteq \nodeLabels$ defining $\ruleGraph$'s \textbf{root}, and a set of \textbf{children} $\ruleChildren$. Each child in $\ruleChildren$ is of the form $(\predicate, \dir, \ruleChild)$ consisting of a predicate $p$ (e.g., \texttt{writtenBy}), the directionality $\dir$ of the rule (i.e., $\dirOut \text{or} \dirIn$), and a descendent rule $\ruleChild$.
A \textbf{leaf} rule has no children, i.e.,  $\ruleGraph_{\text{leaf}} = (\ruleRoot, \emptyset)$.
An \textbf{atomic} rule consists of one root with a single child (e.g., {\small (\texttt{\{Book\}}, \{\texttt{writtenBy}, $\dirIn$, (\texttt{\{Author\}}, $\emptyset$})\})), since all rules can be formed from compositions of these. Rule $g$ in Fig.~\ref{fig:rule-example} (which reads, \texttt{\small Books have fictional family characters and are written by authors who are born in countries.}), rooted at \texttt{\small Book}, consists of three atomic rules, has root $\ruleRoot = \{\small \texttt{Book}\}$ and two children $\ruleChildren$ (for clarity we omit the braces for sets): {\small (\texttt{writtenBy}, $\dirOut$, (\texttt{Author}, ({\texttt{bornIn}, $\dirOut$, ( \texttt{Country}, $\emptyset$)})))} and {\small (\texttt{character}, $\dirIn$, ({\texttt{Fictional Family}}, $\emptyset$))}.

{\subsubsection{\bf Rule Assertion $\assertion_\ruleGraph$} \label{subsec:assertions} An assertion $\assertion_\ruleGraph$ of a rule $\ruleGraph = (\ruleRoot, \ruleChildren)$ over the KG $\graph$ is an instantiation of the edges and labels that $\ruleGraph$ asserts around a particular node, and is reminiscent of a rule \emph{grounding} \cite{meilicke2019anytime}. The set of all assertions of rule $\ruleGraph$ is $\assertions^{(\ruleGraph)}$. Formally, $\assertion_\ruleGraph \in \assertions^{(\ruleGraph)}$ is a subgraph induced by a traversal that starts at a node $\traversalStart \in \nodes$ with at least the same labels as $\ruleRoot$ (i.e., $\ruleRoot \subseteq \labels(\traversalStart)$), and that recursively follows $\ruleGraph$'s syntax. For example, \texttt{\small War \& Peace} is the starting node $\traversalStart$ of one assertion of the rule in Fig.~\ref{fig:rule-example}. If the traversal fails to match the syntax of the rule at any point, then we call it an \textbf{exception} of $\ruleGraph$, in which case the assertion is just the node $\traversalStart \equiv \assertion_\ruleGraph$ that violates the rule. Otherwise the induced subgraph is called a \textbf{correct assertion} of $\ruleGraph$. Formally, $\correctAssertions^{(\ruleGraph)}$ and $\exceptions^{(\ruleGraph)}$ are the set of $\ruleGraph$'s correct assertions and exceptions respectively. Every assertion is either a correct assertion or an exception, so $\correctAssertions^{(\ruleGraph)}$ and $\exceptions^{(\ruleGraph)}$ form a partition of $\assertions^{(\ruleGraph)}$.}

\subsubsection{\bf Minimum Description Length (MDL) Principle} \label{subsec:mdl-def} 
In two-part (\textit{crude})
MDL~\cite{rissanen:78:mdl}, given a set of models $\Models$, the best model $\Model \in \Models$ minimizes $\mdl(\Model) + \mdl(\DB|\Model)$, where
$\mdl(\Model)$ is the length (in bits) of the description of $\Model$, and
$\mdl(\DB|\Model)$ is the length of the description of the data when encoded using $\Model$.
In our work, we leverage MDL to concisely summarize a given KG.

\vspace{0.1cm}
\subsubsection{\bf Problem Definition} Because both errors and missing information are instances of \emph{abnormalities}, we unify KG characterization in terms of what is normal, strange, and missing, as follows: 
\begin{problem}[Inductive KG Summarization]\label{problem:prob2} Given a knowledge graph $\graph$, and an inaccessible ideal knowledge graph $\KGgold$, we seek to find a concise model  $\goldOptimalModel$ of inductive rules that summarize what is normal in both $\graph$ and $\KGgold$. The rules should be (1) interpretable (by which we mean readable in natural language) and, (2) 
their exceptions should reveal abnormal information in the KG, whether it be erroneous (e.g., some $\triple \in \edges : \triple \notin \edgesGold$), missing (e.g., some $\triple \in \edgesGold : \triple \notin \edges$), or a legitimate exception (e.g., some $\triple \in \edges : \triple \in \edgesGold$).
\end{problem}

The \textit{concise} set of rules admits efficient performance on follow-up tasks (such as error detection and incompleteness identification).
Although existing rule mining techniques can be adapted to handle variants of this problem (typically they are tailored to either detect a specific type of error or perform completion), they tend to result in a large number of redundant rules (\S~\ref{subsec:interpret}) and require heuristics to be adapted to tasks that they were not designed for.
In the next section, we formalize our problem definition further and propose a principled, information-theoretic solution that naturally unifies KG characterization. 

\subsection{Inductive Summarization: MDL Model}
\label{sec:mdl}

{The inductive KG summarization problem (Problem~\ref{problem:prob2}) is closely related to the idea of compression in information theory---compression finds patterns in data (what is normal), which in turn can reveal outliers (what is strange or missing). In this work, we leverage MDL (\S~\ref{subsec:mdl-def}) for KG summarization---a formulation that we are the first to use in the context of KG rule mining. Based on our preliminary definitions above, Problem~\ref{problem:prob2} can be restated more formally:}
\begin{problem}[Inductive KG Summarization with MDL] \label{problem:prob3} Given a knowledge graph $\graph$, we seek to find the model $\goldOptimalModel$ (i.e., set of rules) that minimizes the description length of the graph,
\begin{equation}
    \goldOptimalModel = \argmin_{\model \in \modelSpace} L(\graph, \model) = \argmin_{\model \in \modelSpace}\{L(\model) + L(\graph|\model)\},
    \label{eq:mdl}
\end{equation}
where $\model$ is a set of rules (\S~\ref{subsec:model-def}) describing what is normal in $\graph$, $L(\model)$ is the number of bits to describe $\model$, and $L(\graph|\model)$ is the number of bits to describe parts of $\graph$ that $\model$ fails to describe. Thus, expensive parts of $L(\model)$ and $L(\graph|\model)$
reveal abnormal information in $\graph$ (\S~\ref{sec:algo-anomaly}).
\end{problem}

In \S~\ref{subsec:mdl-model} we will define our model space $\modelSpace$, how to describe a KG with a model $\model \in \modelSpace$, and how to encode it in bits. Then, in \S~\ref{subsec:mdl-err} we will describe the KG \emph{under the model}, $L(\graph|\model)$, which we refer to as the model's error, since it encodes what is \emph{not} captured by $\model$. All logarithms are base 2.

\subsubsection{MDL Models $\modelSpace$ and Cost $L(\model)$}
\label{subsec:mdl-model}

A \textbf{model} $\model\in\modelSpace$ is a set of rules, and each rule has a set of correct assertions (or guided traversals of a graph $\graph$, \S~\ref{subsec:assertions}).
The model thus describes $\graph$'s \textit{semantics} (labels) and \textit{connectivity} (edges) through rule-guided traversals over $\graph$. 
Each time a node is visited, some of its labels are revealed by the structure of the rule.
For instance, arriving at the node \texttt{\small Leo Tolstoy} while traversing the subgraph in Fig.~\ref{fig:rule-example}, reveals (i)~its \texttt{\small Author} label, since this is implied by the rule on the left, and (ii)~its link to where the traversal just came from (viz., \texttt{\small War \& Peace}). 

For our model, we consider a classic \textbf{information theoretic transmitter/receiver setting} \cite{shannon1948mathematical}, where the goal is to transmit (or describe) the graph to the receiver using as few bits as possible. In other words, the sender must guide the receiver in how to fill in an empty binary adjacency tensor $\adjacencyTensor$ and binary label matrix $\labMatrix$ with the 1s needed to describe $\graph$. Since MDL seeks to find the best model, the costs that are constant across all models (e.g., the number of nodes and edges) can be ignored during model selection. 
At a high level, beyond this preliminary information, we need to transmit the number of rules (upper bounded by the number of possible \textit{candidate rules}),  
followed by the rules in $\model$ and their assertions, which we discuss in detail next
\begin{equation}
    L(\model) =
    \underbrace{
    \log(2*|\nodeLabels|^2*|\relations|+1)
    }_{\text{\# rules}}
    + \sum_{\ruleGraph \in \rules} \big(
    \underbrace{ L(\ruleGraph) }_{\text{rules}}
    + \underbrace{
    L(\assertions^{(\ruleGraph)})}_{\text{assertions}} \big)
\label{length:model}
\end{equation}

\vspace{0.2cm}
\noindent \textbf{Encoding the Rules.} 
The rules serve as schematic instructions on how to populate the adjacency tensor $\adjacencyTensor$ and label matrix $\labMatrix$ that describe $\graph$. Our rule definition states that a rule $\ruleGraph = (\ruleRoot, \ruleChildren)$ consists of a set of root labels
$\ruleRoot$ (semantics) and recursive rule definitions of the children $(\predicate, \dir, \ruleChild) \in \ruleChildren$ (structure), so we need to transmit both of them to the receiver. We encode them as
\begin{equation}
\resizebox{0.92\hsize}{!}{
    $L(\ruleGraph) =
    \underbrace{L(\ruleRoot)}_{\text{root labels}}
    + \underbrace{\naturalEncoding(|\ruleChildren| + 1)}_{\text{\# children}}
    + \displaystyle \sum_{\ruleChild \in \ruleChildren}\big(
    - \underbrace{\log\frac{n_\predicate}{\numEdges}}_{\text{predicate}} +
    \smallunderbrace{\mystrut{0.35ex}  1}_{\text{dir}} +
    \underbrace{L(\ruleChild)}_{\text{child rule}} \big)$,
    }
    \label{length:rule}
\end{equation}
where $n_\predicate$ is the number of times predicate $\predicate$ occurs in $\graph$. We discuss each term in Eq.~\eqref{length:rule} next.
    We encode the \textit{root labels} $\ruleRoot$ by transmitting their number (upper bounded by $\numNodeLabels$) and then the actual labels via optimal prefix code~\cite{cover2012elements}, since they may not occur with the same frequency:
\begin{equation}
    L(\ruleRoot) =
    \underbrace{\mystrut{1ex}  \log|\nodeLabels|}_{\text{\# labels}} + \underbrace{\textstyle \sum_{\ell \in \ruleRoot} -\log\frac{n_{\ell}}{\numNodes}}_{\text{labels}},
    \label{length:rule-node-labels}
\end{equation}
where $n_{\ell}$ is the number of times label $\ell \in \nodeLabels$ occurs in $\graph$. Then, for the \textit{children} $\ruleChildren$, we transmit their number (expected to be small) using the optimal encoding of an unbounded natural number~\cite{rissanen:83:integers} similarly to \cite{akoglu2013mining} and denoted $\naturalEncoding$; and per child we specify:
(i)~its predicate $p$ using an optimal prefix code as in Eq.~\eqref{length:rule-node-labels},
(ii)~its directionality $\delta$ (i.e., $\dirOut$ or $\dirIn$) without making an \emph{a priori} assumption about which is more likely,
and (iii)~its descendent rule $\ruleChild$, by recursively applying Eq.~\eqref{length:rule} until leaf rules (with 0 children) are reached.

We note that while some labels can be inferred from rules (e.g., the \texttt{\small Author} label of \texttt{\small Leo Tolstoy}), it is possible that all labels will not be revealed by rules. Thus, we transmit the un-revealed labels as \emph{negative error}---i.e., information needed to make the transmission lossless, but that is \emph{not} modeled by $\model$. We discuss this in \S~\ref{subsec:mdl-err}.

So far, in our running example, once the receiver has the information that \texttt{\small War \& Peace} is a book, it can apply the rule in Fig.~\ref{fig:rule-example}. It knows that \texttt{\small War \& Peace} should have one or more \texttt{\small Fictional Families} as characters, and one or more \texttt{\small Authors} who wrote it, but it does not yet know \emph{which} \texttt{\small Fictional Families} and \texttt{\small Authors}. This information will be encoded next in the assertions.

\vspace{0.2cm}
\noindent \textbf{Encoding the Rule Assertions.}
In Eq.~\eqref{length:model}, the last term encodes the assertions, $\assertions^{(\ruleGraph)}$, of each rule $\ruleGraph$.
The receiver infers the starting nodes of the traversals
from $\ruleGraph$’s root (Eq.~\eqref{length:rule}) and the node labels (encoded via other rules or $\negativeLabelError$ in Eq.~\eqref{length:err-labels}). Thus, we transmit the failed traversals (i.e., exceptions) and details needed to guide the correct assertions:

\begin{equation}
    L(\assertions^{(\ruleGraph)}) = \underbrace{L(\exceptions^{(\ruleGraph)})}_{\text{exceptions}} + \underbrace{L(\correctAssertions^{(\ruleGraph)}) \vphantom{\exceptions^{(\ruleGraph)}} }_{\text{correct assertions}},
    \label{length:rule-assertions}
\end{equation}
\noindent
The first term transmits which assertions are \textbf{exceptions} to a rule 
(e.g. the book \texttt{\small Syntactic Structures}, which is non-fiction and hence does not have any \texttt{\small Fictional Family} characters). 
We transmit the number of exceptions, followed by their IDs (i.e., which assertions they are), chosen from among the assertions:
\begin{equation}
   L(\exceptions^{(\ruleGraph)}) = \underbrace{\mystrut{2.1ex} \log|\assertions^{(\ruleGraph)}|}_{\text{\# exceptions}}
  + \underbrace{\log {|\assertions^{(\ruleGraph)}| \choose |\exceptions^{(\ruleGraph)}| }}_{\text{exception ids}},
  \label{length:rule-exceptions}
\end{equation}
where $\log|\assertions^{(\ruleGraph)}|$ is an upper bound on the number of exceptions.

Intuitively, the bits needed to encode exceptions penalize overly complex rules, which are unlikely to be accurate and generalizeable.

The remaining traversals are correct assertions, for
which we transmit details as we traverse
each $\assertion_\ruleGraph \in \correctAssertions^{(\ruleGraph)}$. The encoding cost for $\correctAssertions^{(\ruleGraph)}$ is the sum of the cost of all these traversals:
\begin{equation}
    L(\correctAssertions^{(\ruleGraph)}) = 
    \textstyle \sum_{\assertion_\ruleGraph \in \correctAssertions^{(\ruleGraph)}}
    L(\assertion_\ruleGraph). 
    \label{length:rule-correct-assertions}
\end{equation}
Each traversal is encoded by recursively
visiting neighbors according to the
recursive structure of $\ruleGraph$. Formally,

\begin{equation}
    L(\assertion_\ruleGraph) =
    \sum_{\ruleChild \in \ruleChildren}
    \big(
    \underbrace{\mystrut{3.5ex} \log \numNodes}_{\text{\# neighbors}}
    + \underbrace{\mystrut{3.5ex} \log{ \numNodes - 1 \choose |\correctAssertions^{(\ruleChild)}| }}_{\text{neighbor ids}}
    + \underbrace{\sum_{\assertion_{\ruleChild} \in \correctAssertions^{(\ruleChild)}}
    L(\assertion_{\ruleChild})}_{\text{proceed recursively}}
    \big),
    \label{length:traversal}
\end{equation}
where, for each child of $\ruleGraph$, we first transmit the number of $\assertion_\ruleGraph$'s neighbors with the child's labels (upper-bounded by the number of nodes $\numNodes$ in $\graph$), followed by the neighbors' IDs (which are the starting nodes of the child rule's correct assertions, since the child is itself a rule) using a binomial transmission scheme. Once the neighbors have been revealed, the traversal proceeds recursively to them. For example, the traversal in Fig.~\ref{fig:rule-example} begins at \warAndPeace{} and the rule has two children (\texttt{\small characters} and \texttt{\small authors}). For each, we transmit the number of nodes relevant (5 and 1 respectively), followed by their IDs. The traversal then proceeds recursively to each node just specified.

\vspace{0.15cm}
\subsubsection{MDL Error $L(\graph|\model)$}
\label{subsec:mdl-err}

In Eq.~\eqref{eq:mdl}, along with sending the model $\model$, we also need to send anything not modeled, i.e., the model's \emph{negative error}. This error consists of the cost of encoding (i) the node labels that are not revealed by the rules and (ii) the unmodeled edges. We denote the modeled labels and edges as $\explainedLabels$ and $\explainedEdges$ respectively, which contain the subset of 1s in $\adjacencyTensor$ and $\labMatrix$ that the receiver has been able to fill in via the rules it received in $\model$. 
We denote the unmodeled labels and edges as the binary matrix $\negativeLabelError = \labMatrix - \explainedLabels$ and binary tensor $\negativeEdgeError = \adjacencyTensor - \explainedEdges$, and these are what we refer to as \emph{negative error}. The cost of the model's error is thus
\begin{equation}
    L(\graph|\model) = L(\negativeLabelError) + L(\negativeEdgeError).
    \label{length:err}
\end{equation}

\noindent
Specifically, the receiver can infer the number of missing node labels (i.e., 1s in $\negativeLabelError$)
given the total number of node labels and the number not already explained by the model (\S~\ref{subsec:mdl-model}).
Thus, we send only the \emph{position} of the 1s in $\negativeLabelError$, encoding over a binomial (where $|.|$ denotes set cardinality and the number of 1s in a tensor/matrix):
\begin{equation}
    L(\negativeLabelError)
    = \log{|\nodeLabels|\cdot\numNodes - |\explainedLabels| \choose |\negativeLabelError|},
    \label{length:err-labels}
\end{equation}
We transmit missing edges $L(\negativeEdgeError)$ analogously
\begin{equation}
    L(\negativeEdgeError) =
    \log{\numNodes^2 \cdot |\relations| - |\explainedEdges| \choose |\negativeEdgeError|}.
    \label{length:err-edges}
\end{equation}

\section{Inductive Summarization: Method}
\label{sec:algorithm}

In the previous section, we fully defined the encoding cost $L(\graph,\model)$ of a knowledge graph $\graph$ with a model $\model$ of rules. Here we introduce our  method, \method, which will leverage our KG encoding $L(\graph, \model)$ to find a concise summary $\goldOptimalModel$ of inductive rules, with which it will characterize what is normal, what is strange, and what is missing in the KG.

A necessary step to this end is to generate a set of candidate rules $\candidates \supseteq \goldOptimalModel$ from which MDL will construct the rules that can best compress $G$.
However, even given that set,
selecting the optimal model $\goldOptimalModel\in\modelSpace$ involves a combinatorial search space, since any subset of $\candidates$ is a valid model, i.e., $|\modelSpace| = 2^{|\candidates|}$ (where even $|\candidates|$ 
can be in the millions for large KGs). This cannot be searched exhaustively, and our MDL search space does not have easily exploitable structure, such as the anti-monotonicity property of the support/confidence framework.
To find a tractable solution, we exploit the compositionality of rules---starting with simple, atomic rules and building from there. 
We give \method's pseudocode in Alg.~\ref{alg:alg}, and describe it by line next. 

\subsection{Generating and Ranking Candidate Rules}
\label{subsec:candidates}
\subsubsection{Candidate Generation (line 1).}\label{subsec:rulegen} We begin by generating \emph{atomic} candidate rules---those that assert exactly one thing (\S~\ref{subsec:rule}). 
The number of possible atomic rules is exponential in the number of node labels, but
not all of them need to be generated: rules that never apply in $G$ 
do not explain any of the KG, and hence will not be selected by MDL.
Thus, we use the graph to guide candidate generation.
For each edge in the graph, \method generates atomic rules that could explain it.
For instance, the edge \tolstoy{}
could be explained by rules such as, ``\texttt{\small books are written by authors}'' and ``\texttt{\small authors write books}.'' These have the forms $\ruleGraph_1 = (\{\texttt{\small Book}\}, \{\texttt{\small writtenBy}, \dirOut, (\{\texttt{\small Author}\}, \emptyset)\})$ and $\ruleGraph_2 = (\{\texttt{\small Author}\}, \{\texttt{\small writtenBy}, \dirIn, (\{\texttt{\small Book}\}, \emptyset)\})$, respectively. To avoid candidate explosion from allowing rules to have any subset of node labels, we only generate atomic rules with a single label per node here, and account for more combinations of labels in the next step.

\subsubsection{Qualifying Candidate Rules with Labels (line 2).} \label{subsec:qualifiers} Adding more labels to rules can help make them more accurate and more inductive by limiting the number of places they apply (e.g., Fig.~\ref{fig:real-rule}), and subsequently their exceptions (which incur a cost in our MDL model). 
To this end, given a rule $\ruleGraph$, \method{} identifies the labels shared by all the starting nodes of the correct assertions of the rule: $\RootLabelSet = \bigcap_{\assertion_\ruleGraph \in \correctAssertions^{(\ruleGraph)}} \labels(\traversalStart)$. If this set contains more labels than the rule (i.e.,  
$\ruleRoot \subset \RootLabelSet$),
then it forms a new rule $\ruleGraph'$ with root $\RootLabelSet$. 
If $L(\graph,\emptyModel \cup \{\ruleGraph'\}) \leq L(\graph,\emptyModel \cup \{\ruleGraph\})$, where $\emptyModel$ is the empty model without any rules, \method{} replaces $\ruleGraph$ with $\ruleGraph'$ in $\candidates$.
It carries this out for all the rules in 
$\candidates$. This can be viewed as \emph{qualifying} $\ruleGraph$: it qualifies the conditions under which $\ruleGraph$ applies, to those that contain \emph{all} the labels rather than its original label alone.

\enlargethispage{\baselineskip}

\subsubsection{Ranking Candidate Rules (line 3).}
\label{subsec:ranking} 
Considering all possible combinations of candidate rules $\powerset({\candidates})$, and finding the optimal model $\goldOptimalModel$ is not tractable.
Moreover, an alternative greedy approach that constructs the model by selecting in each iteration the rule $\ruleGraph \in \candidates$ that leads to the greatest reduction in the encoding cost, would still be quadratic in $|\candidates|$ 
(which is in the order of many millions for large-scale KGs).
Instead, for scalability, given the set of (potentially qualified) candidate rules $\candidates$, we devise a ranking that allows \method to take a constant number of passes over the candidate rules.
Intuitively, our ranking considers the amount of explanatory power that a rule has---i.e., how much reduction in error it could lead to: 
\begin{equation}
    \Delta L(\graph|\emptyModel \cup \{\ruleGraph\}) = L(\graph|\emptyModel) - L(\graph|\emptyModel \cup \{\ruleGraph\}).
    \label{eq:rank}
\end{equation}
\method{} sorts the rules descending on this value, and breaks ties by considering rules with more correct assertions first. If that fails, the final tie-breaker is the lexicographic ordering of rules' root labels.

\begin{algorithm}[t!]
	\caption{\method}
	\label{alg:alg}
	\begin{algorithmic}[1]
	    \small
		\Statex \textbf{Input}: Knowledge graph $\graph$ 
		\Statex \textbf{Output}: A model $\Model$, consisting of a set of rules
		\State Read $\graph$ and generate candidate rules $\candidates$ \Comment{\textcolor{gray}{\S~\ref{subsec:rulegen}}}
		\State Qualify candidate rules with labels 
		\State Rank all rules $\ruleGraph \in \candidates$ first by $\downarrow \Delta L(\graph|\emptyModel)$
		then by $\downarrow$ $|\correctAssertions(\ruleGraph)|$ and  $\downarrow$ lexicographic $\ruleRoot$ \Comment{\textcolor{gray}{\S~\ref{subsec:ranking}}, Eq.~\eqref{eq:rank}}
		\State $\model \gets \emptyset$
		\While{\text{not converged}} \Comment{\textcolor{gray}{i.e., more rules can be added to $\model$}}
		    \For{$\ruleGraph \in \candidates$}
		        \If{$L(\graph,\model \cup \{g\}) < L(\graph, \model)$} \Comment{\textcolor{gray}{\S~\ref{subsec:selection}}}
                 \State $\model \gets \model \cup \{g\}$
                 \State $\candidates \gets \candidates\setminus \{\ruleGraph\}$
		        \EndIf
            \EndFor
		\EndWhile
		\State Optionally perform refinements \RefTwo{} and \RefThree{} \Comment{\textcolor{gray}{\S~\ref{subsec:refinements}}}
	\end{algorithmic}
\end{algorithm}

\subsection{Selecting and Refining Rules}
\label{subsec:algo-mdl}

\subsubsection{Selecting Rules (lines 4-9).} \label{subsec:selection} 
After ranking the candidate rules $\candidates$, \method{} initializes $\model = \emptyset$ and considers each $\ruleGraph \in \candidates$ in ranked order for inclusion in $\model$. For each rule $\ruleGraph$, it computes $L(\graph, \model \cup \{\ruleGraph\})$, i.e., the MDL objective if $\ruleGraph$ is added to the current model. If this is less than the MDL cost $L(\graph, \model)$ without the rule (e.g., rule \ruleGraph\ correctly explains new parts of \graph), then \method adds $\ruleGraph$ to $\model.$ If $\ruleGraph$ has a reverse version (e.g., ``\texttt{\small books are written by authors}'' and ``\texttt{\small authors write books}''), \method considers both at once and picks the one that gives a lower MDL cost. \method{} runs a small number of passes over $\candidates$ until no new rules are added. The resulting model $\model$ approximates the true optimal model $\goldOptimalModel$.

\subsubsection{Refining Rules (line 10).} \label{subsec:refinements}
The model at this point only contains atomic rules.
To better approximate $\goldOptimalModel$, we introduce two refinements that compose rules via merging \RefTwo{} and nesting \RefThree{}. 

\textbf{Refinement \RefTwo{} for ``\textit{rule merging}''} composes rules that share a \emph{root}. It identifies all sets of rules, $\{\ruleGraph_i, \dots, \ruleGraph_j\}$ with matching roots that correctly apply in the same cases, i.e., $\mathcal{L}_{\ruleGraph_i} = \dots = \mathcal{L}_{\ruleGraph_j}$ and $\{s_{\assertion_{\ruleGraph_i}} : \assertion_{\ruleGraph_i} \in \correctAssertions^{(\ruleGraph_i)}\} = \dots = \{s_{\assertion_{\ruleGraph_j}} : \assertion_{\ruleGraph_j} \in \correctAssertions^{(\ruleGraph_j)} \}$. It then merges these into a single rule $\ruleGraph'$, consisting of the union of the children ${\ruleChildren}_i \cup \dots \cup{\ruleChildren}_j$. For example, if all books that have authors ($\ruleGraph_1$) also have publishers ($\ruleGraph_2$), then these would be merged into a single rule. We refer to this variant as {\bf \methodTwo{}}.

\textbf{Refinement \RefThree{} for ``\textit{rule nesting}''} considers composing rules where an inner node of one rule $\ruleGraph_{in}$ matches the root of another rule $\ruleGraph_{rt}$, possibly creating a more inductive rule. 
{\RefThree{} begins by computing, between each compatible $\ruleGraph_{in}$ and $\ruleGraph_{rt}$, the Jaccard similarity of the correct assertion starting points of the matching inner and root nodes (i.e., it quantifies the `fit' of the nodes). For instance, if a rule asserts that ``\texttt{\small books have authors}'' $(\ruleGraph_{in})$, and another rule asserts that ``\texttt{\small authors have a birthplace}'' ($\ruleGraph_{rt}$), then the Jaccard similarity is computed over the set of book authors and the set of authors with birthplaces.
The refinement then considers nesting the rules in descending order of Jaccard similarity, resulting in rule $\ruleGraph_{rt}$ being subsumed into rule $\ruleGraph_{in}$, which becomes its ancestor.
If the composed rule $\ruleGraph_{in} \circ \ruleGraph_{rt}$ leads to lower encoding cost than the individual rules (e.g., $\ruleGraph_{in}$ qualifies $\ruleGraph_{rt}$ as in Fig.~\ref{fig:real-rule}),  i.e., $L(\graph,(\model \setminus \{\ruleGraph_{in}, \ruleGraph_{rt}\})\cup \{\ruleGraph_{in} \circ \ruleGraph_{rt}\}) < L(\graph, \model)$, then the composition replaces $\ruleGraph_{in}$ and $\ruleGraph_{rt}$.
The Jaccard similarity between rules that were compatible with $\ruleGraph_{in}$ or $\ruleGraph_{rt}$ is then recomputed with $\ruleGraph_{in} \circ \ruleGraph_{rt}$, the list of compatible rules
is re-sorted by Jaccard similarity, and the search continues until all pairs are considered with none being composed (i.e., when no composition reduces the encoding cost). }
This sorting is done over the set of selected rules 
$\model$ (\S~\ref{subsec:selection}), where $|\model| << |\candidates|$, and since only few compositions occur 
(\S~\ref{subsec:compression}), this repeated sorting is tractable. As nesting embeds $\ruleGraph_{rt}$ into $\ruleGraph_{in}$, this refinement allows for arbitrarily expressive rules to form. We call our method with both refinements (merging and nesting) 
{\bf \methodThree{}}.

\subsection{Deriving Anomaly Scores}
\label{sec:algo-anomaly}
We now discuss how to leverage a model $\model$ (i.e., a summary of rules) mined by \method{} towards identifying what is strange or anomalous in a KG, whether it be erroneous or missing---two key tasks in KG research. Anomaly detection seeks to identify objects that differ from the norm \cite{akoglu2015graph,Aggarwal16_outlier}. In our case, the learned summary concisely describes \textit{what is normal} in a KG.

Intuitively, nodes that \emph{violate} rules, and edges that are \emph{unexplained} are likely to be anomalous. Next, we make this intuition more principled by
defining anomaly scores for entities (nodes) and relationships (edges) in information theoretic terms.

\subsubsection{Entity Anomalies} \label{subsec:anomaly-node}
We define the anomaly score $\anom$ of an \emph{entity} or \emph{node} $v$ as the number of bits needed to describe it as an exception to the rules in the model:
\begin{equation}
\anom(v) = \sum_{\substack{\ruleGraph \in r(v)\ :\\ v \in \exceptions^{(\ruleGraph)}}}
\underbrace{\frac{1}{|\exceptions^{(\ruleGraph)}|}\log {|\assertions^{(\ruleGraph)}| \choose |\exceptions^{(\ruleGraph)}| }}_{\text{bits to model $v$ as an exception}},
\label{eq:node-anomaly-score}
\end{equation}\noindent
where $r: \nodes \rightarrow \powerset(\rules)$ 
maps each node to the rules that apply to it. We distribute the cost of the exceptions equally over all $|\exceptions^{(\ruleGraph)}|$ violating nodes, following Eq.~\eqref{length:rule-exceptions}.

\subsubsection{Relationship Anomalies}
We also introduce an anomaly score for a  \emph{relationship} or \emph{edge}. 
Intuitively, edges that are not explained by the model $\model$ are anomalous, and their anomaly score is defined as the number of bits describing them as negative error.
Since we transmit all unmodeled edges in $\negativeEdgeError$ together (Eq.~\eqref{length:err-edges}), we make this intuition more principled by distributing this transmission cost evenly across all unmodeled edges in the anomaly score: 
\begin{equation}
\anom^{(p)}(s, p, o) =
\begin{cases}
      \underbrace{\frac{1}{|\negativeEdgeError|}\log{\numNodes^2 * |\relations| - |\explainedEdges| \choose |\negativeEdgeError|}}_{\text{bits spent transmitting edge as neg err}} & \text{if } \negativeEdgeErrorElement = 1  \\
      0 & \text{otherwise}
  \end{cases}
\label{eq:edge-anom-score-1}
\end{equation}
Under the reasonable assumption that our model effectively captures what is \emph{normal} in the KG, it follows that edges unexplained by the model are likely to be \emph{abnormal}. Equation~\eqref{eq:edge-anom-score-1} captures this notion, but to prevent
the unexplained edges from all receiving equal scores, we add the anomaly score of the endpoints (Eq. \eqref{eq:node-anomaly-score}):
\begin{equation}
\anom(s, p, o) =
    \anom(s)+\anom(o)+\anom^{(p)}(s, p, o).
\label{eq:edge-anom-score-2}
\end{equation}

\subsection{Complexity Analysis} Generating candidate rules involves iterating over each edge (and its nodes' labels) as it is encountered. 
The number of possible atomic rules with a single label that could explain an edge $(s, p, o)$ is $2\cdot|\labels(s)|\cdot|\labels(o)|$. Letting $\phi_{max}$ denote the max number of labels over all nodes, 
the overall complexity of candidate generation is $\order(\numEdges\cdot\phi_{max}^2)$.
The number of candidate rules generated, $|\candidates|$, is also $\order(\numEdges\cdot\phi_{max}^2)$.
Computing the error, $L(\graph|\model)$, is constant since it only involves computing the log-binomials.
The computation of $L(\model)$ 
depends on the time of traversing and describing the correct assertions. Since the traversals occur in a DFS manner (in linear time) over 
a subgraph enough smaller than $\graph$ to be ignored, $L(\model)$ takes $\order(|\model|)$ time. 
Since ranking only requires computing $L(\graph|\model)$, which is a small constant, the cost comes only from sorting $|\candidates|$ items, which is $\order(|\candidates|\log|\candidates|)$.
\method{} takes a small number of passes over the candidate set (\S~\ref{subsec:selection}) in $\order(|\candidates|)$ time.
So, the overall complexity is $\order(|\candidates| + |\candidates|\log|\candidates|)$, which simplifies to $\order(|\candidates|\log|\candidates|)$, or $\order(\numEdges \cdot \phi_{max}^2 \log(\numEdges \cdot \phi_{max}^2))$. We omit the complexity of the refinements for brevity.

\section{Evaluation}
\label{sec:experiments}
\label{sec:eval}

\begin{table}[b!]
\centering
\footnotesize
\vspace{-0.35cm}
\caption{Description of KG datasets: number of nodes, edges, node labels, relations, and average / median labels per node, resp.}
\label{table:stats}
\vspace{-0.35cm}
\begin{tabular}{l r r r r r r r}
\toprule
  & $\mathcal{|V|}$ & $\mathcal{|E|}$ & $\mathcal{|L_V|}$ & $\mathcal{|L_E|}$ & avg $\labels(v)$ & med $\labels(v)$ \\
\midrule
NELL & 46,682 & 231,634 & 266 & 821 & 1.53 & 1 \\
DBpedia & 976,404 & 2,862,489 & 239 & 504 & 2.72 & 3 \\
Yago & 6,349,336 & 12,027,848 & 629,681 & 33 & 3.81 & 3 \\
\bottomrule
\end{tabular}
\vspace{-0.4cm}
\end{table}

Our experiments seek to answer the following questions:
\begin{itemize}
    \item[{\bf \qOne{}.}] Does \method{} characterize what is normal? How well can \method{} compress, in an interpretable way, a variety of KGs? 
    \item[{\bf \qTwo{}.}] Does \method{} identify what is strange? Can it identify and characterize multiple types of errors?
    \item[{\bf \qThree{}.}] Does \method{} identify what is missing?
    \item[{\bf \qFour{}.}] Is \method{} scalable?
\end{itemize}

\begin{table*}[ht]
\begin{minipage}[t]{0.62\linewidth}
\centering
\footnotesize
	\caption{Compression: The small \% bits needed (relative to an empty model) and number of rules found by various models demonstrate the effectiveness of \method{} variants at finding a concise set of rules in $\graph$.
	\amie{}~\cite{galarraga2015fast} finds Horn rules, which
	cannot be encoded with our model, so we only report the number of rules it finds.
	\freq{} and \cover{} are baseline models that we introduce
	by greedily selecting from our candidate set $\candidates$ (\emph{without} MDL) the top-$k$ rules that (1)~ correctly apply the most often and (2)~ cover the most edges, resp.
	For these, we preset $k$ to the number of rules found by the best-compressed version of our method and report it as top-$k$ to distinguish from the non-preset values.
     }
	\label{table:compression}
	\vspace{-0.2cm}
	\begin{tabular}{p{1.0cm}lccccccc}
		\toprule
		& & {Horn rules} && \multicolumn{5}{c}{Rules of the form $g=(\ruleRoot, \ruleChildren)$} \\ \cmidrule{3-3} \cmidrule{5-9}
		\emph{Dataset} & \emph{Metric} & \amie{}  && \freq{} & \cover{}  & \kgistNull{} & \methodTwo{} & \methodThree{}\\
		\toprule
		\multirow{3}{*}{  {\parbox{1.1cm}{\NELL{} \textcolor{gray}{(6,268,200 bits)}}} }
        & \% Bits needed & \textcolor{gray}{N/A}  && 191.46\%  & 192.72\%  & 73.88\%  &73.00\%  & \cellcolor{gray!15}63.57\%  \\
        & Edges Explained & \textcolor{gray}{N/A} && 57.33\%  & 50.12\%  & \cellcolor{gray!15}78.52\%  & \cellcolor{gray!15}78.52\%  & 74.67\% \\
        & \# Rules & 32,676  && top-$k$ & top-$k$  & 1,115 & 647 & \cellcolor{gray!15}573\\
        \cmidrule{1-9}
        \multirow{3}{*}{ {\parbox{1.1cm}{\DBpedia{} \textcolor{gray}{(119,117,468 bits)}}} }
        & \% Bits needed  & \textcolor{gray}{N/A}  && 674.51\% & 718.22\%  & 69.88\% & 69.84\% & \cellcolor{gray!15}69.77\%\\
        & Edges Explained & \textcolor{gray}{N/A}  && 80.64\% & 71.70\%  & \cellcolor{gray!15}89.17\% & \cellcolor{gray!15}89.17\% & 88.51\% \\
        & \# Rules & $\sim$6,963 \cite{galarraga2015fast} && top-$k$ & top-$k$ & 516 & 505 & \cellcolor{gray!15}498\\
        \cmidrule{1-9}
        \multirow{3}{*}{ {\parbox{1.1cm}{\Yago{} \textcolor{gray}{(793,027,801 bits)}}} }
        & \% Bits needed & \textcolor{gray}{N/A}  && 896.33\% & 947.64\% & 76.13\% & 75.98\% & \cellcolor{gray!15}75.04\% \\
        & Edges Explained & \textcolor{gray}{N/A}  && 86.54\% & 83.44\% & \cellcolor{gray!15}88.40\% & \cellcolor{gray!15}88.40\% & 85.20\%\\
        & \# Rules & failed  && top-$k$ & top-$k$ & 60,298 & 34,331 & \cellcolor{gray!15}32,670\\
  	    \bottomrule
	\end{tabular}
\end{minipage}
\hfill
\begin{minipage}[t]{0.35\linewidth}
    \vspace{-0.4cm} 
    \centering
    \captionof{figure}{Rules mined from \NELL{} (left two) and \DBpedia{} (right). 
    While the bottom atomic rule in Rule 3 does not hold in general (not all places are the river mouth of bodies of water), qualifying (\S~\ref{subsec:qualifiers}) \& \RefThree{} (\S~\ref{subsec:refinements}) improve its inductiveness: since rules apply to the root (black node), the bottom rule is ``qualified'' to only apply to those places that are tributaries of \texttt{\footnotesize Places, Streams, \& Bodies of Water}.}
    \label{fig:real-rule}
    \includegraphics[width=\columnwidth]{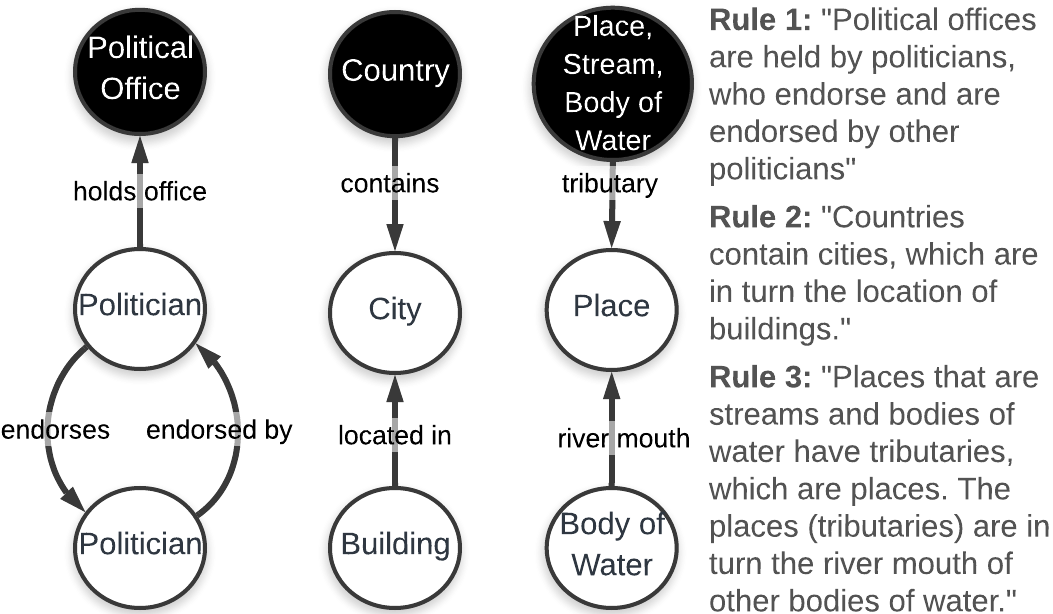}
\end{minipage}
\end{table*}

\noindent \textbf{Data.} Table~\ref{table:stats} gives descriptive statistics for our data:
\NELL{} \cite{carlson2010toward} or ``Never-Ending Language Learning'' continually learns facts via crawling the web. 
Our version contains 1,115 iterations, each introducing new facts for which the confidence has grown sufficiently large. 
\DBpedia{} \cite{auer2007dbpedia} is extracted from Wikipedia data, heavily using the structure of infoboxes. The extracted content is aligned with the \DBpedia{} ontology via crowd-sourcing \cite{paulheim2017knowledge}.
\Yago{} \cite{suchanek2007yago}, like \DBpedia{}, is built largely from Wikipedia. 
\Yago{} contains 3 orders of magnitude more node labels than the other two graphs (Tab.~\ref{table:stats}).

\subsection{[\qOne] What is normal in a KG?} 
\label{subsec:compression}
In this section, we demonstrate how \method{} characterizes what is \emph{normal} in a KG
by achieving (1) high compression, (2)~concise, and (3)~interpretable summaries with intuitive rules.

\subsubsection{KG Compressibility.} Although compression is \textit{not} our goal, it is our \textit{means} to evaluate the quality of the discovered rules. Effective compression means that the discovered rules describe the KG accurately and concisely.
\label{subsec:compression}

\vspace{0.15cm}
\noindent \textbf{Setup.} We run \method{} on all three KGs since each has different properties (Tab.~\ref{table:stats}). $\emptyModel$ denotes an empty model with no rules, corresponding to transmitting the graph entirely as error, i.e., $L(\graph, \emptyModel) = L(\graph|\emptyModel)$. We compare compression over this model.

\vspace{0.15cm}
\noindent \textbf{Baselines.} We compare to: (i) {\bf \freq{}} which, instead of using MDL to select rules from $\candidates$, selects the top-$k$ rules that correctly apply the most often, where we set $k$ to be the number selected by the best compressed version of \method{}. (ii) {\bf \cover{}} is directly analogous to \freq{}, replacing the metric of frequency with the number of edges explained by the rule. Both select rules independently, without regard for whether rules explain the same edges. (iii) {\bf \amie{}}~\cite{galarraga2015fast} finds Horn rules, which cannot be encoded with our model, so we do not report compression results, but only the number of rules it finds. While other KG compression techniques exist (\S~\ref{subsec:graph-sum}), we are seeking to find inductive rules that are useful for refinement, whereas generic graph compression methods compress the graph, but never generate rules, and are hence not comparable.

\vspace{0.15cm}
\noindent
\textbf{Metrics.} 
For each dataset, the first row reports the percentage of bits needed for $L(\graph, \model)$ relative to the empty model. That is, it reports $L(\graph, \model) / L(\graph, \model_0)$. Small values occur when $\model$ compresses $\graph$ well, and hence smaller values are better. 
The second row reports the percentage of edges explained: $|\explainedEdges| / |\adjacencyTensor|$. Lastly, we report how many rules were selected to achieve the results.

\vspace{0.15cm}
\noindent\textbf{Results.} We record KG compression in bits in Table~\ref{table:compression}. In all cases, \method{} is significantly more effective than the \freq{} and \cover{} baselines, which ignore MDL. Indeed, \freq{} and \cover{} result in values greater than 100\% in the first row, meaning they lead to an \emph{increase} in encoding cost over $\emptyModel$, due in part to selecting rules independently from each other, and hence potentially explaining the same parts of the graph with multiple rules. \method{} is very effective at explaining the graph, leaving only a small percentage of the edges unexplained.
It also explains more edges than \cover{} due to rule overlap again.
The two refinements, \RefTwo{} and \RefThree{}, are also effective at refining model $\model$ to more concisely describe $\graph$. \RefThree{}, which allows arbitrarily expressive rules, refines $\model$ to contain fewer
and better compressing rules. \methodThree{} explains slightly fewer edges than \methodTwo{} because nested rules apply only when their root does (e.g., Fig \ref{fig:real-rule}).

\enlargethispage{\baselineskip}

\vspace{-0.05cm}
\subsubsection{Rule Conciseness \& Interpretability}
\label{subsec:interpret}
We compare the number of rules mined by \method to that of \amie{} \cite{galarraga2015fast}. For \amie{}, we set min-support to 100 and min PCA confidence to 0.1, as suggested 
by the authors \cite{galarraga2015fast}. When running \amie{} on graphs larger than \NELL{}, we experienced intolerable runtimes (inconsistent with those in~\cite{galarraga2015fast}). For \Yago{} we were unable to get results, while for \DBpedia{} we report numbers from \cite{galarraga2013amie} on an older, but similarly sized version of \DBpedia{}.
In Tab.~\ref{table:compression} we see that \method{} mines orders of magnitude fewer rules than \amie{}, showing that it is more computationally tractable to apply our concise summary of rules
to refinement tasks than the sheer number of rules obtained by other rule-mining methods that operate in a support/confidence framework. This is because redundant rules cost additional bits to describe, so MDL encourages conciseness. While these other methods could use the min-support parameter to reduce the number of rules, it is not clear how to set this parameter \emph{a priori}. Using MDL, we can approximate the optimal number of rules in a parameter-free, information-theoretic way, leading to fewer but descriptive rules. 

Furthermore, we present and discuss in Fig.~\ref{fig:real-rule} example rules mined with \method{}. These show that our rules are interpretable and intuitively inductive, and that \RefThree{} is a useful refinement for improving the inductiveness of rules.

\subsection{[\qTwo] What is strange in a KG?} 
\label{subsec:anomaly}
Here we quantitatively analyze the effectiveness of \method{} at identifying a diverse set of anomalies, and demonstrate the interpretability of what it finds. Whereas most approaches focus on
exceptional facts \cite{zhang2018maverick}, erroneous links, erroneous node type information \cite{paulheim2017knowledge}, or identification of incomplete information (e.g., link prediction) \cite{galarraga2015fast}, \method{} rules can be used to address multiple of these at once. To evaluate this, we inject anomalies of multiple types into a KG, and see how well \method{} identifies them.

\begin{table*}[ht]
\begin{minipage}[t]{1.0\linewidth}
\centering
\footnotesize
	\caption{Anomaly detection results on \NELL{}.
    We mark the best performing method with a gray background and the best \emph{unsupervised} method with bold text.
    We mark statistical significance at a 0.05 $p$-value (paired t-test) with an ``*'' for
	\method{}\_\freq{}/\methodTwo{} vs. unsupervised methods.
	The final row shows the average rank of each method.
	\methodTwo{} performs the most consistently well. 
	}
	\label{table:anomaly-detection}
	\begin{tabular}{p{1.1cm}lccc@{}ccccc}
		\toprule
		& & \multicolumn{2}{c}{Supervised} && \multicolumn{4}{c}{Unsupervised} \\ \cline{3-4} \cline{6-9}
		\emph{Task} & \emph{Metric} & \complex & \transe{} && \sdv{} & \amie{} & \kgistFreq{} & \methodTwo{}\\
		\toprule
		\multirow{4}{*}{\parbox{1.1cm}{\centering {\em All anomalies}}}
		& AUC & 0.5508 $\pm$ 0.02 & 0.5779 $\pm$ 0.04 && 0.4996 $\pm$ 0.00 & 0.4871 $\pm$ 0.04 & 0.5739 $\pm$ 0.01 & \textbf{\cellcolor{gray!15}0.6052 $\pm$ 0.03}* \\
        & P@100 & 0.4820 $\pm$ 0.05 & 0.7040 $\pm$ 0.06 && 0.5100 $\pm$ 0.04 & 0.3980 $\pm$ 0.07 & 0.6816 $\pm$ 0.10 & \textbf{\cellcolor{gray!15}0.7419 $\pm$ 0.07}* \\
        & R@100 & 0.0087 $\pm$ 0.00 & 0.0126 $\pm$ 0.00 && 0.0092 $\pm$ 0.00 & 0.0072 $\pm$ 0.00 & 0.0126 $\pm$ 0.00 & \textbf{\cellcolor{gray!15}0.0139 $\pm$ 0.00}* \\
        & F1@100 & 0.0172 $\pm$ 0.00 & 0.0247 $\pm$ 0.00 && 0.0181 $\pm$ 0.00 & 0.0141 $\pm$ 0.00 & 0.0247 $\pm$ 0.01 & \textbf{\cellcolor{gray!15}0.0273 $\pm$ 0.01}* \\
        \cmidrule{1-9}
        \multirow{4}{*}{\parbox{1.1cm}{\centering {\em \anomOne{} missing labels}}}
        & AUC & 0.5842 $\pm$ 0.04 & 0.6021 $\pm$ 0.06 && 0.4997 $\pm$ 0.00 & 0.4409 $\pm$ 0.06 & 0.5149 $\pm$ 0.02 & \textbf{\cellcolor{gray!15}0.6076 $\pm$ 0.03}* \\
        & P@100 & 0.2640 $\pm$ 0.05 & 0.4280 $\pm$ 0.15 && 0.3040 $\pm$ 0.06 & 0.1200 $\pm$ 0.05 & 0.4067 $\pm$ 0.11 & \textbf{\cellcolor{gray!15}0.4759 $\pm$ 0.05}* \\
        & R@100 & 0.0119 $\pm$ 0.00 & 0.0181 $\pm$ 0.01 && 0.0134 $\pm$ 0.00 & 0.0057 $\pm$ 0.00 & 0.0199 $\pm$ 0.01 & \textbf{\cellcolor{gray!15}0.0244 $\pm$ 0.01}* \\
        & F1@100 & 0.0227 $\pm$ 0.01 & 0.0346 $\pm$ 0.01 && 0.0257 $\pm$ 0.01 & 0.0109 $\pm$ 0.01 & 0.0377 $\pm$ 0.01 & \textbf{\cellcolor{gray!15}0.0463 $\pm$ 0.02}* \\
        \cmidrule{1-9}
        \multirow{4}{*}{\parbox{1.15cm}{\centering {\em \anomTwo{} superfluous labels}}}
        & AUC & 0.5502 $\pm$ 0.02 & \cellcolor{gray!15}0.5659 $\pm$ 0.03 && 0.4989 $\pm$ 0.01 & 0.4946 $\pm$ 0.03 & 0.4997 $\pm$ 0.04 & \textbf{0.5115 $\pm$ 0.03} \\
        & P@100 & 0.1780 $\pm$ 0.05 & \cellcolor{gray!15}0.3160 $\pm$ 0.16 && 0.2160 $\pm$ 0.07 & 0.1040 $\pm$ 0.09 & 0.2081 $\pm$ 0.06 & \textbf{0.2485 $\pm$ 0.09} \\
        & R@100 & 0.0122 $\pm$ 0.00 & \cellcolor{gray!15}0.0219 $\pm$ 0.01 && 0.0152 $\pm$ 0.00 & 0.0070 $\pm$ 0.01 & 0.0169 $\pm$ 0.01 & \textbf{0.0175 $\pm$ 0.01} \\
        & F1@100 & 0.0229 $\pm$ 0.00 & \cellcolor{gray!15}0.0408 $\pm$ 0.02 && 0.0283 $\pm$ 0.01 & 0.0131 $\pm$ 0.01 & 0.0311 $\pm$ 0.01 & \textbf{0.0326 $\pm$ 0.01} \\
        \cmidrule{1-9}
        \multirow{4}{*}{\parbox{1.1cm}{\centering {\em \anomThree{} erroneous links}}}
        & AUC & 0.2495 $\pm$ 0.03 & 0.4126 $\pm$ 0.08 && 0.4966 $\pm$ 0.01 & \textbf{\cellcolor{gray!15}0.8902 $\pm$ 0.08} & 0.7383 $\pm$ 0.00 & 0.8423 $\pm$ 0.00 \\
        & P@100 & 0.1020 $\pm$ 0.04 & 0.0020 $\pm$ 0.00 && 0.0480 $\pm$ 0.02 & \textbf{\cellcolor{gray!15}0.1860 $\pm$ 0.08}* & 0.0131 $\pm$ 0.01 & 0.0137 $\pm$ 0.01 \\
        & R@100 & 0.0374 $\pm$ 0.02 & 0.0007 $\pm$ 0.00 && 0.0176 $\pm$ 0.01 & \textbf{\cellcolor{gray!15}0.0679 $\pm$ 0.03}* & 0.0051 $\pm$ 0.01 & 0.0052 $\pm$ 0.01 \\
        & F1@100 & 0.0548 $\pm$ 0.02 & 0.0011 $\pm$ 0.00 && 0.0257 $\pm$ 0.01 & \textbf{\cellcolor{gray!15}0.0995 $\pm$ 0.05}* & 0.0074 $\pm$ 0.01 & 0.0075 $\pm$ 0.01 \\
        \cmidrule{1-9}
        \multirow{4}{*}{\parbox{1.0cm}{\centering {\em \anomFour{} swapped labels}}}
        & AUC & 0.5369 $\pm$ 0.03 & 0.5527 $\pm$ 0.02 && 0.4991 $\pm$ 0.00 & 0.4891 $\pm$ 0.03 & \textbf{\cellcolor{gray!15}0.6904 $\pm$ 0.01}* & 0.6633 $\pm$ 0.07 \\
        & P@100 & 0.2160 $\pm$ 0.08 & 0.4200 $\pm$ 0.09 && 0.2080 $\pm$ 0.08 & 0.1240 $\pm$ 0.06 & \textbf{\cellcolor{gray!15}0.5360 $\pm$ 0.15}* & 0.4768 $\pm$ 0.10 \\
        & R@100 & 0.0136 $\pm$ 0.00 & 0.0269 $\pm$ 0.01 && 0.0128 $\pm$ 0.00 & 0.0079 $\pm$ 0.00 & \textbf{\cellcolor{gray!15}0.0379 $\pm$ 0.01}* & 0.0320 $\pm$ 0.01 \\
        & F1@100 & 0.0256 $\pm$ 0.01 & 0.0505 $\pm$ 0.01 && 0.0241 $\pm$ 0.01 & 0.0148 $\pm$ 0.01 & \textbf{\cellcolor{gray!15}0.0705 $\pm$ 0.01}* & 0.0599 $\pm$ 0.01 \\
        \midrule
        \parbox{1.15cm}{\centering \textbf{Avg rank}} & & 4.10 & 2.90 && 4.15 & 5.00 & 2.90 & \cellcolor{gray!15}\textbf{1.95}\\
  	    \bottomrule
	\end{tabular}
\end{minipage}
\end{table*}

\vspace{0.15cm}
\noindent {\bf Setup.} We inject four types of anomalies.
For each,
we select $\samplingProb$ percent of $\graph$'s nodes uniformly at random to perturb. We sample nodes independently for each type, so it is possible that occasionally a node is chosen multiple times. This is realistic, since there are multiple types of errors in KGs at once \cite{paulheim2017knowledge}.
Although we target nodes, our perturbations also affect their incident edges. Thus, we formulate the anomaly detection problem as identifying the perturbed edges.
Specifically, we introduce the following anomalies:
\begin{itemize*}
\item  \anomOne{} {\em Missing labels}: We remove one label from each node.
    Unlike the 
    \anomTwo{}-\anomFour{},
    we only sample nodes with more than one label. E.g., we may remove the \texttt{\small entrepreneur} label from \texttt{\small Bill Gates}, leaving the labels
    \texttt{\small billionaire}, etc. We consider all the in/out edges of the altered nodes as perturbed.

\item \anomTwo{} {\em Superfluous labels}: We add to each node a randomly selected label that it does not currently have. E.g., we may add the label \texttt{\small Fruit} to \texttt{\small Taj Mahal}.

\item \anomThree{} {\em Erroneous links}: We inject 1 or 2 edges incident to each node, choosing the edge's predicate and destination randomly. E.g., we may
 inject random edges like \texttt{\small (Des Moines, owner, Coca-Cola)}. We mark injected edges as anomalous.

\item  \anomFour{} {\em Swapped labels}: For each node, we replace a label with a new random one that it does not yet have. 
\end{itemize*}
For this experiment we show results on \NELL{}, since it has confidence values for each of its edges, which we can use to sample negative examples. The perturbed edges are ground truth errors (positive examples), and we randomly sample from \NELL{} an equal number of ground truth correct edges with a confidence value of 1.0 (after filtering out edges that our injected anomalies perturbed). 
We use a 20/80 validation/test split, and the perturbed graph for training.

\vspace{0.15cm}
\noindent {\bf Baselines.} We compare to (i)~ \textbf{\complex{}}, an embedding method that we tune as in~\cite{trouillon2016complex} (ranking edges based on its scoring function), (ii)~ \textbf{\transe{}}, an embedding method that we tune as in~\cite{bordes2013translating} (ranking edges based on their energy scores), (iii)~ \textbf{\sdv{}}~\cite{paulheim2014improving}, an error detection method deployed in \DBpedia{} (it outputs an edge ranking), and (iv)~ \textbf{\amie{}}, designed for link prediction, but which we adapt for error detection by ranking based on the sum of the confidences of the rules that predict each test edge (i.e., edges that are predicted by many, high-confidence rules will be low in the ranking, and edges that are not predicted by any rules will be high in the ranking). We also tried PaTyBRED \cite{melo2017detection}, but it had prohibitive runtime. 

\vspace{0.15cm}
\noindent \textbf{\method{} variants.} To define edge anomalies for our variants, we use the edge-based anomaly score $\anom$ in Eq.~\eqref{eq:edge-anom-score-2}. 
\kgistFreq is the \freq{} method described in \S~\ref{subsec:compression}, but uses \method{}'s anomaly scores. While \methodThree{} learns compositional rules that help with compression, we found that the simpler rules of \methodTwo{} performed better in this task, so we report only its results for brevity. The unsupervised methods do not have hyper-parameters, but are tested on the same test set as \complex{}/\transe{}, so the validation set errors are additional noise they must overcome.

\vspace{0.15cm}
\noindent {\bf Metrics.}
Each ranking includes only the test set edges, and we compute 
the AUC for each ranking, using reciprocal rank as the predicted score for each edge---edges higher in the ranking being closer to 1 (i.e., more anomalous). We also compute Precision@100, Recall@100, and F1@100 for (i)~ the entire test set of edges; (ii)~ each type of perturbed edges from the different anomaly types. For ties in the ranking, we extend the list beyond 100 until the tie is broken (e.g., if the 100th and 101st edge have the same score, then we compute over 101 edges). 
Ties did not often extend much beyond 100, but for \kgistFreq they tended to extend farthest. 
Positives are considered perturbed edges and negatives un-perturbed edges. When computing scores
for
a particular anomaly type, we first filter the ranking to only contain the edges perturbed by \emph{that} anomaly type and the un-perturbed edges to ensure that edges perturbed by other anomaly types are not considered false negatives.

\begin{figure}[t!]
    \vspace{-0.3cm} 
    \captionsetup{type=figure}
    \centering
    \caption{Example anomalies in \NELL{} (left two) and \DBpedia{} (right) that violate many rules.
    The 1st and 3rd are
    missing information.
    While most states are the headquarters of sports teams, Pennsylvania does not have any teams listed. However, 
    the  Steelers, Eagles, etc. are all teams located
    in PA. 
    Also, unlike most music genres, 4-beat has no persons listed who play it. The 2nd exception may
    not capture missing
    information, since hippopotamuses were, until recently, considered herbivores; this node is anomalous as it differs from many carnivorous and omnivorous mammals.
    }
    \includegraphics[width=.7\linewidth]{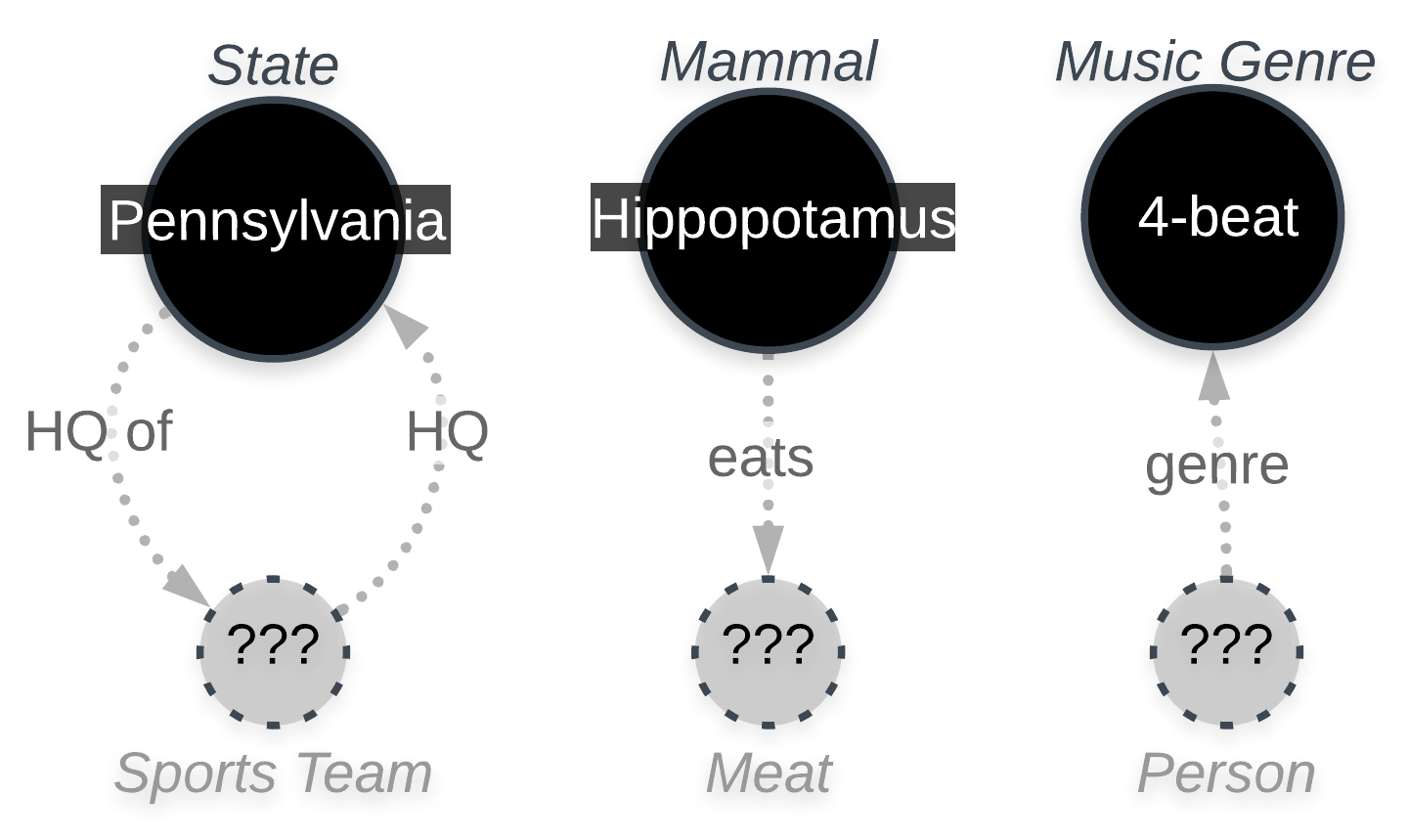}
    \label{fig:anomaly-examples}
    \vspace{-0.3cm}
\end{figure}

\vspace{0.15cm}
\noindent {\bf Results.} In Table \ref{table:anomaly-detection} we report results identifying anomalies generated with sampling probability $\samplingProb = 0.5\%$ and 5 random seeds. We report avg and stdev over the 5 perturbed graphs. Across all anomaly types, \methodTwo{} is most effective at identifying anomalous edges, demonstrating its generality. This is further evidenced by its top average ranking: it ranks 1.95 on average across all anomaly types and metrics.
Furthermore, as discussed in Fig.~\ref{fig:anomaly-examples}, not only can it identify anomalies, but its interpretable rules allow us to reason about \emph{why} something is anomalous.

In most cases, \methodTwo{} even outperformed \complex{} and \transe{}, supervised methods requiring validation data for hyper-parameter tuning. \anomTwo{} is the only anomaly type where supervised methods outperform unsupervised methods, but the difference is not statistically significant. \kgistFreq performs better than most other baselines, demonstrating that our formulation of anomaly scores and rules are effective at finding anomalies. However, as \methodTwo{} usually outperforms \kgistFreq, we conclude that MDL leads to improvement over simpler rule selection approaches. \amie{} only performed well on \anomThree{}. We conjecture that this is because randomly injected edges are likely to be left un-predicted by \emph{all} of \amie{}'s rules. On the other hand, edges with perturbed endpoints may be left un-predicted by \emph{some} rules, but, out of the large number of rules that \amie{} mines (\S~\ref{subsec:interpret}), \emph{some} rule is likely to still predict it. The results for $\samplingProb = 1.0\%$ were overall consistent, with a few fluctuations between \methodTwo{} and \kgistFreq. We omit the results for brevity.

\subsection{[\qThree] What is missing in a KG?} 
\label{subsec:kgc}
In this section, we evaluate \method{}'s ability to find missing information. Most KG completion methods target link prediction, which seeks to find missing links between pairs of nodes that are \emph{present in} a KG. If either node
is missing, then link prediction cannot provide any information.
We focus on this task: revealing \emph{where} entities are missing. Since \method{}'s rules apply to \emph{nodes}, rather than \emph{edges}, the rule exceptions can reveal \emph{where} links to seen \emph{or unseen} entities are missing (but cannot predict which specific entity the link should be to). Thus, our task and link prediction are complementary.

\vspace{0.15cm}
\noindent {\bf Setup.} 
We assume the commonly used partial completeness assumption (PCA) \cite{paulheim2017knowledge, galarraga2013amie, galarraga2017predicting}, according to which, if an entity has one relation of a particular type, then it has \emph{all} relations of that type in the KG (e.g., a \texttt{\small movie} with at least one \texttt{\small actor} listed in the KG has all its \texttt{\small actors} listed).
We generate a \textit{perturbed KG} with ground-truth incomplete information via the following steps: (1)
we randomly remove $\samplingProb\%$ of nodes (and their adjacent edges) from $\graph$,
and (2) we enforce the PCA (e.g., if we removed one \texttt{\small actor} from a \texttt{\small movie}, then we remove all the \texttt{\small movie}'s \texttt{\small actor} \emph{edges}).
Our goal is to identify that the neighbors of the removed nodes are missing information, and what that information is. We run \method{} on the perturbed KG, 
and identify the exceptions, $\exceptions^{(\ruleGraph)}$, of each rule $\ruleGraph \in \model$. If a rule asserts the removed information, then this is a true positive. For example, if we removed \frankenstein's \texttt{\small author} and \method mines the rule that \texttt{\small \texttt{\small books} are written by \texttt{\small authors}}, then that rule asserts the removed information. We use \NELL{} and \DBpedia{} for this experiment as their sizes permit several runs over different perturbed KG variants. 

\vspace{0.15cm}
\noindent {\bf Baselines.} Link prediction methods are typically used for KG completion, but they do not apply to our setting: they require that \emph{both} endpoints of an edge be in $\graph$ to predict the edge, while our setup assumes that one endpoint is \textit{missing} from $\graph$.
Thus, we compare \method{} to \freq{} and \amieC{}. \freq{}, as before (\S~\ref{subsec:compression}), selects the top-$k$ rules with the most correct assertions, where $k$ is set to the number of rules \method{} mines. \amieC{} is what we name the method from \cite{galarraga2017predicting}. \amieC{} requires training data comprised of examples of ($u$, \texttt{\small incomplete}, $p$) triples where $u \in \nodes$ is an entity, $p \in \relations$ is a predicate, and the triple specifies that node $u$ is missing its links of type $p$ (e.g., a \texttt{\small movie} is missing \texttt{\small actors}). We use 80\% of the removed data as training data for \amieC{} and test all methods on the remaining 20\%. We tune \amieC{}'s parameters as in \cite{galarraga2017predicting}. 

\vspace{0.15cm}
\noindent {\bf Metrics.} We report only recall, $\mathbf{R}$, since information that we did not remove but was reported missing could be either a false positive, or real missing information that we did not create \cite{paulheim2017knowledge}. We compute recall as the number of nodes identified as missing, divided by the total number of nodes removed. In addition, we compute a more strict recall variant, $\mathbf{R_L}$, which requires that the missing node's \emph{label} also be correctly identified (e.g., not only do we need to predict the absence of a missing \texttt{\small writtenBy} edge, but also that the edge should be connected to an \texttt{\small author} node). \method can return this label information, but \amieC{} only predicts the missing link, not the 
label. Thus, we omit {$\mathbf{R_L}$} for \amieC{}.

\vspace{0.15cm}
\noindent {\bf Results.}
We report results in Table~\ref{table:incompleteness} with $\samplingProb = 5\%$ (the results are consistent with other $\samplingProb$ values). \method{} outperforms all baselines by a statistically significant amount (10-11\% on $\mathbf{R}$, 13-27\% on $\mathbf{R_L}$, and paired t-test $p$-values $<< 0.01$), which demonstrates its effectiveness in finding information missing from a KG. We conjecture that \method{} outperforms \amieC{} because \amieC{} requires all training data be focused around a small number of predicates (e.g., 10-11 in \cite{galarraga2017predicting}) in order to learn effective rules, while MDL encourages \method{} to explain as much of the KG as possible, allowing it to find missing information over more diverse regions of the KG. This is further evidenced by the fact that \method{} outperforms \freq{}, which only applies to frequent regions of the KG. Not only does \method{} effectively reveal where the missing nodes are, it also usually correctly identifies their labels (small drop in $\mathbf{R_L}$ compared to $\mathbf{R}$). \amieC{} is not able to do this, and \freq{} can only report the label sometimes, by taking advantage of the rules we formulated in this work.

\begin{table}[t!]
    \centering
    \caption{KG completion results on \NELL{} and \DBpedia{} (averages and stdevs of 10 runs). We report Recall (R) and Recall + Label ($\mathbf{R_L}$), a stronger version of R that requires both the location of the missing node \emph{and} its label be identified.
    We list link prediction~(LP) methods to emphasize that our task and LP are complimentary.
    Results are statistically significant (paired t-test) with $p$-values $<< 0.01$.}
    \label{table:incompleteness}
    \vspace{-0.35cm}
    \resizebox{\columnwidth}{!}{ 
    \begin{tabular}{p{1.0cm}lccccc}
		\toprule
		& & \multicolumn{2}{c}{Supervised} && \multicolumn{2}{c}{Unsupervised} \\
		\cmidrule{3-4} \cmidrule{6-7}
		\emph{Dataset} & \emph{Metric} & LP & \amieC{} \cite{galarraga2017predicting} && \freq{} & \method{} \\
		\toprule
		\multirow{2}{*}{ {\parbox{1.1cm}{ \NELL{} }} }
		& $\mathbf{R}$ & \textcolor{gray}{N/A} & $0.6587 \pm 0.03$&& $0.4589 \pm 0.02$ & \cellcolor{gray!15}$0.7598 \pm 0.02$ \\
		& $\mathbf{R_L}$ & \textcolor{gray}{N/A} & \textcolor{gray}{N/A} && $0.3924 \pm 0.02$ & \cellcolor{gray!15}$0.6636 \pm 0.01$\\
        \cmidrule{1-7}
        \multirow{2}{*}{ {\parbox{1.1cm}{ \DBpedia{} }} }
        & $\mathbf{R}$ & \textcolor{gray}{N/A} & $0.8187 \pm 0.01$ && $0.8049 \pm 0.01$ & \cellcolor{gray!15}$0.9288 \pm 0.00$\\
        & $\mathbf{R_L}$ & \textcolor{gray}{N/A} & \textcolor{gray}{N/A} && $0.7839 \pm 0.01$ & \cellcolor{gray!15}$0.9179 \pm 0.00$\\
  	    \bottomrule
	\end{tabular}
	}
\end{table}

\subsection{\bf Scalability} \label{subsec:scale}

\setlength{\columnsep}{5pt}
\begin{wrapfigure}{r}{0.5\linewidth}
    \centering
    \includegraphics[width=0.75\linewidth]{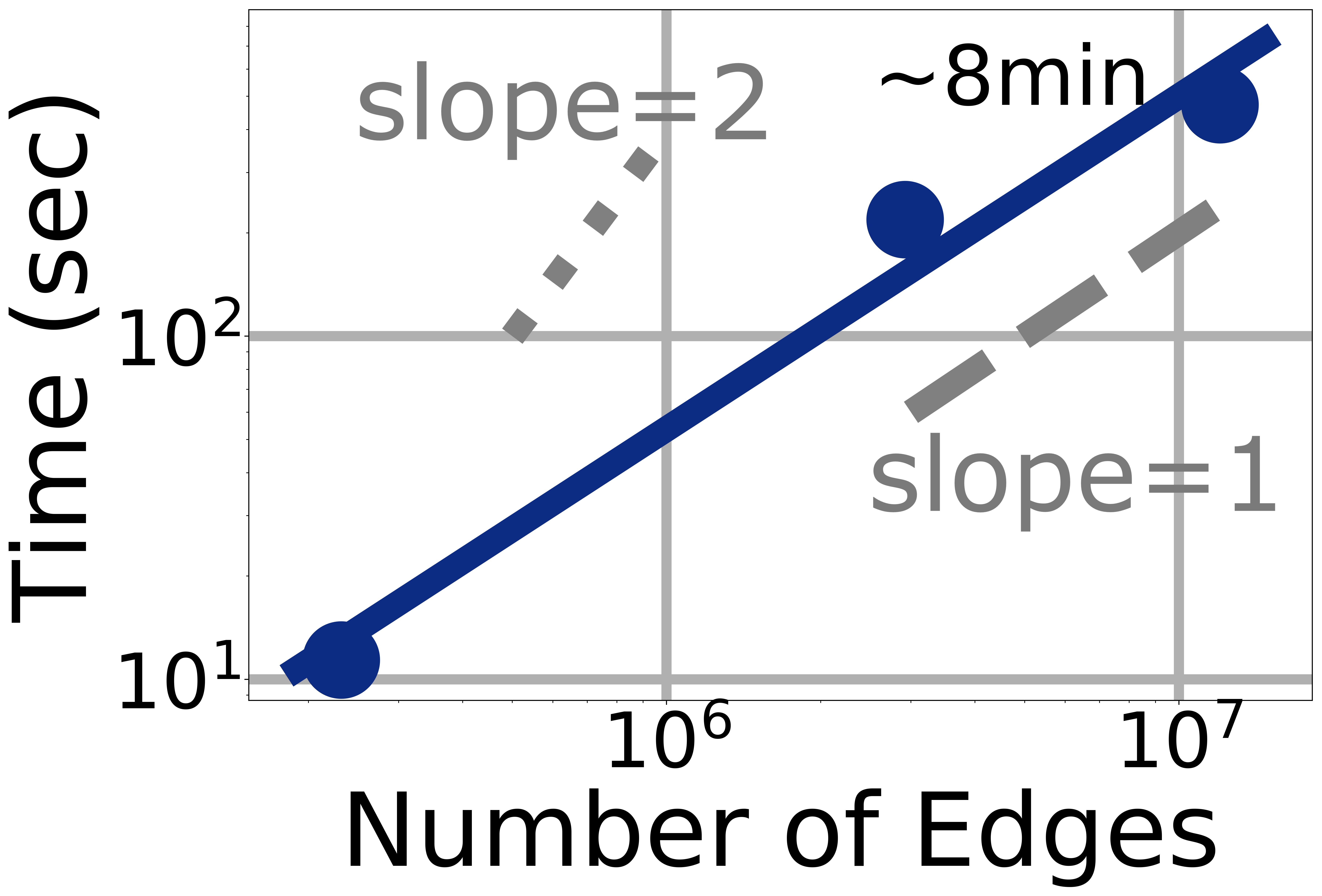}
    \vspace{-0.2cm}
    \caption{\method{} is near-linear in the number of edges.}
    \label{fig:runtime-all}
    \vspace{-0.3cm}
\end{wrapfigure}
In this section, we evaluate \method{}'s performance as the number of edges in the KG grows. We perform this evaluation on an Intel(R) Xeon(R) CPU E5-2697 v3 @ 2.60GHz with 1TB RAM, using a Python implementation.
\NELL{}, \DBpedia{}, and \Yago{}
have from 231,634 to 12,027,848 edges. 
We run \method{} on each KG three times. Since we aim to analyze the runtime with respect to the number of edges, but \Yago{} has three orders of magnitude more labels, we run \method{} with an optimization that only generates candidate rules with the 300 most frequent labels (approximately equal to \NELL{} and \DBpedia{}), allowing us to fairly investigate the effect of the number of \emph{edges}.
Figure \ref{fig:runtime-all} shows the number of edges vs. runtime in seconds. In practice, \method{} is near-linear in the number of edges. Even on \Yago{}, it mines summaries in only a few minutes, and on \NELL{} in seconds.

\section{Conclusion}
\label{sec:conclusion}
This paper proposes a unified, information theoretic approach to KG characterization, \method{}, which solves our proposed problem of inductive summarization with MDL. \method{} describes what is \emph{normal} in a KG with a set of interpretable, inductive rules, which we define in a new, graph-theoretic way. The rule exceptions, and the parts of the KG that the summary fails to describe reveal what is \emph{strange} and \emph{missing} in the KG. \method{} detects various anomaly types and incomplete information in a principled, unified manner, while scaling nearly linearly with the number of edges in a KG---this property allows it to be applied to large, real-world KGs. Future work could explore using \method{}'s rules to guide KG construction.

\vspace{-0.05cm}
\section*{Acknowledgements}
{
This material is based upon work supported by the National Science Foundation under Grant No. IIS 1845491, Army Young Investigator Award No. W911NF1810397, an Adobe Digital Experience research faculty award, and an Amazon faculty award.
Any opinions, findings, and conclusions or recommendations expressed in this material are those of the author(s) and do not necessarily reflect the views of the National Science Foundation or other funding parties.
}

\balance
\bibliographystyle{plain}
\bibliography{abbreviations,References,all}

\end{document}